\documentclass[lettersize,journal]{IEEEtran}
\usepackage{amsmath,amsfonts}
\usepackage{algorithmic}
\usepackage{algorithm}
\usepackage{array}
\usepackage[caption=false,font=normalsize,labelfont=sf,textfont=sf]{subfig}
\usepackage{textcomp}
\usepackage{stfloats}
\usepackage{url}
\usepackage{verbatim}
\usepackage{graphicx}
\usepackage{cite}
\usepackage{times}
\usepackage{epsfig}
\usepackage{amssymb}
\usepackage{booktabs}
\usepackage{pifont}
\usepackage{colortbl}
\usepackage{xcolor}
\usepackage{color}
\usepackage{marvosym}
\usepackage{tikz}

\begin{document}

\title{SRCD: Semantic Reasoning with Compound Domains for Single-Domain Generalized Object Detection}

\author{Zhijie Rao, 
Jingcai Guo\textsuperscript{\Letter},~\IEEEmembership{Member, IEEE},
Luyao Tang,
Yue Huang\textsuperscript{\Letter},\\
Xinghao Ding,
and Song Guo,~\IEEEmembership{Fellow, IEEE}
%
\thanks{Z. Rao, L. Tang, Y. Huang, and X. Ding are with the School of Information Science and Engineering, Xiamen University, Xiamen 361005, China (e-mail: raozhijie@stu.xmu.edu.cn; lytang@stu.xmu.edu.cn; huangyue05@gmail.com; dxh@xmu.edu.cn).}
\thanks{J. Guo and S. Guo are with Department of Computing, The Hong Kong Polytechnic University, Hong Kong SAR., China, and with The Hong Kong Polytechnic University Shenzhen Research Institute, Shenzhen 518057, China (e-mail: jc-jingcai.guo@polyu.edu.hk; song.guo@polyu.edu.hk).}
\thanks{\textit{Corresponding authors: Jingcai Guo and Yue Huang.}}
}

\markboth{Manuscript}%
{Rao \MakeLowercase{\textit{et al.}}: {SRCD: Semantic Reasoning with Compound Domains for Single-Domain Generalized Object Detection}
}

\maketitle

\begin{abstract}
This paper provides a novel framework for single-domain generalized object detection (i.e., Single-DGOD), where we are interested in learning and maintaining the semantic structures of self-augmented compound cross-domain samples to enhance the model's generalization ability. Different from DGOD trained on multiple source domains, Single-DGOD is far more challenging to generalize well to multiple target domains with only one single source domain. 
Existing methods mostly adopt a similar treatment from DGOD to learn domain-invariant features by decoupling or compressing the semantic space. 
However, there may have two potential limitations: 1) pseudo attribute-label correlation, due to extremely scarce single-domain data; and 2) the semantic structural information is usually ignored, i.e., we found the affinities of instance-level semantic relations in samples are crucial to model generalization. 
In this paper, we introduce \underline{S}emantic \underline{R}easoning with \underline{C}ompound \underline{D}omains (SRCD) for Single-DGOD. Specifically, our SRCD contains two main components, namely, the texture-based self-augmentation (TBSA) module, and the local-global semantic reasoning (LGSR) module. TBSA aims to eliminate the effects of irrelevant attributes associated with labels, such as light, shadow, color, etc., at the image level by a light-yet-efficient self-augmentation. Moreover, LGSR is used to further model the semantic relationships on instance features to uncover and maintain the intrinsic semantic structures. 
Extensive experiments on multiple benchmarks demonstrate the effectiveness of the proposed SRCD.
\end{abstract}


\section{Introduction}

\begin{figure}
    \centering
    \begin{minipage}{\linewidth}
    \includegraphics[width=0.99\textwidth]{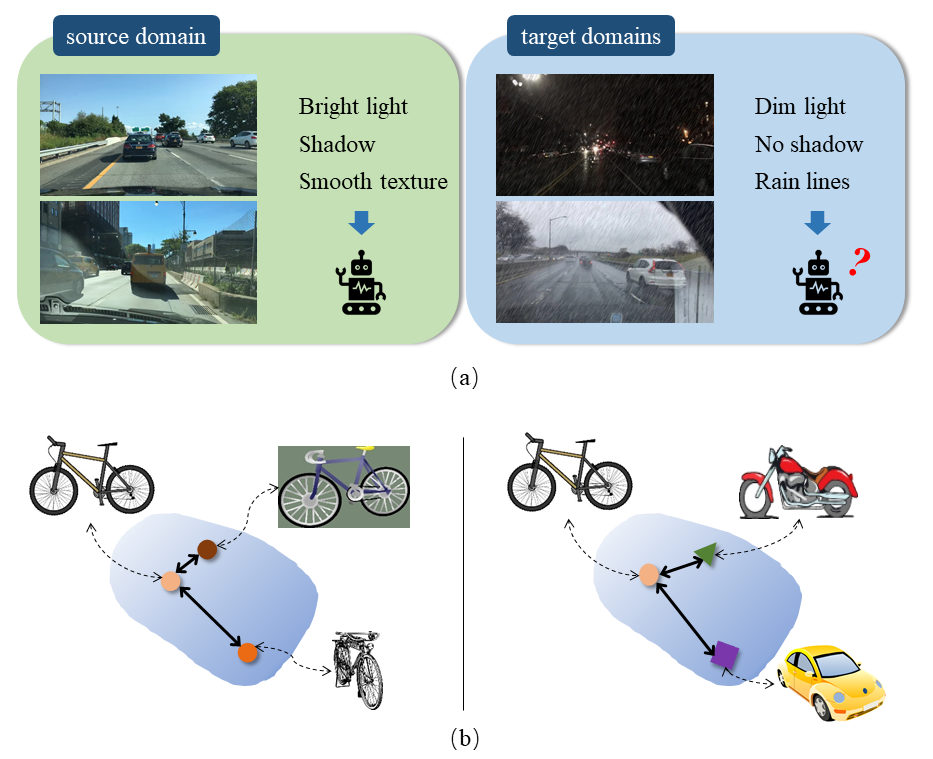}
    \end{minipage}
    \caption{(a) The source domain-specific features may be treated as the domain-invariant information and thus introduce bias into the learned model. (b) Intra- and inter-class logical relationships exist inherently among the samples, which are of crucial importance to improve the model's generalization ability.}
    \label{fig:xiaotu1}
\end{figure}

\IEEEPARstart{O}{bject} detection aims to identify and localize the target of interest in the scene. Previous detection techniques based on deep convolution networks have made tremendous progress in the past few years~\cite{faster2015towards}. 
However, existing off-the-shelf detectors still suffer from distribution discrepancy, i.e., domain shift, that is caused by common factors such as different weathers, regions, or styles. When the training and testing data are not independent identically distributed (non-i.i.d.), 
the performance of the detector may degrade dramatically~\cite{ben2010theory, chen2018domain}. 
In this regard, a series of research has been conducted on how to suppress the effect of domain shift and learn a generalized model. 

Among them, unsupervised domain adaptation (UDA) is a widely investigated direction that learns to relieve the impact of domain shift for object detection, which attempts to transfer shared knowledge from a labeled source domain to an unlabeled target domain. Although some works have obtained several promising results~\cite{chen2018domain, zhao2020collaborative, saito2019strong, shen2019scl, wu2021vector, xu2020exploring, chen2020harmonizing}, UDA relies heavily on the strong assumption that the target domain is accessible during training, which is hardly to be satisfied nor applicable in real-world scenarios. 
To address this issue, domain generalization (DG) follows a more practical setting of cross-domain learning without access to the target domain~\cite{li2018domainMMD, zhou2020deep,zhou2021mixstyle,li2022uncertainty,zhou2022domain}. Specifically, the goal of DG-based object detection (DGOD) is to train a sufficiently generalized model on several available source domains and evaluate it directly on the target domain. 
However, existing DGOD methods mostly require the support of multiple source domains, which is usually costly and time-consuming in data collection and annotation. 
In recent years, some research considers a more challenging and practical problem setup in DGOD, where a sufficiently generalized and robust detection model is trained with only one single source domain, namely, single-DGOD. 
Existing Single-DGOD methods can be categorized into three mainstreams including feature regularization~\cite{pan2018two, pan2019switchable, huang2019iterative, choi2021robustnet}, feature decoupling~\cite{wu2022single}, and consistency constraints~\cite{zhao2022shade}, all of which share a similar treatment from conventional DGOD that regularize and encourage the models to learn domain-invariant feature representation. 

Despite the simplicity and effectiveness of existing off-the-shelf detectors, such methods overlook two critical issues that may limit the generalization ability of learned models, and in turn, degrade the detection performance:  
\ding{182} First, since only one source domain is available during training, the extremely scarce data may bring about some pseudo attribute-label correlations among samples. For example, vehicles on sunny days are usually accompanied by shadows, which may not appear on rainy days (Fig.~\ref{fig:xiaotu1} (a)). 
In this regard, the learned models may tend to adopt the source domain-specific features together with real domain-invariant features to construct the detector, and therefore, introduce significant bias to the learned models. 
\ding{183} Worse still, we found that the semantic structural information in samples is usually ignored. Notably, we observe that the affinities of instance-level semantic relations in samples are of crucial importance to the model generalization ability. 
Such semantic relationships are reflected in intra- and inter-classes (Fig.~\ref{fig:xiaotu1} (b)). For example, bicycles with the same view should be semantically closer than those with different views. In contrast, bicycles and motorcycles are quite similar to each other, and thus they should be semantically closer than any pair with cars. 
Such a structural relationship does not change with domain changes. Recent studies~\cite{chenmix} have also shown that maintaining the semantic structures inherently between samples can facilitate the cross-domain reasoning ability of the model.

To address the above issues, we propose a novel single-DGOD framework that utilizes semantic reasoning with compound domains, termed SRCD, to learn generalized representation by uncovering and maintaining the semantic relation with self-augmented compound domains. 
Specifically, our method consists of two components including texture-based self-augmentation (TBSA) and local-global semantic reasoning (LGSR). 
The goal of TBSA is to eliminate the interference of low-order information such as textures for category determination, i.e., pseudo-association between irrelevant attributes and labels. To achieve this goal, TBSA grafts the style of local patches to the whole image, changing the image style while preserving the semantic information to avoid overfitting the model to the source domain. 
Meanwhile, grey level co-occurrence matrix (GLCM)~\cite{haralick1973textural} is introduced to evaluate the texture complexity of the patterns to filter out valuable patches. 
TBSA provides abundant style-diverse augmented samples, converting the single source domain into compound domains. 
Moreover, the goal of LGSR is to uncover and learn the latent semantic structures from the compound domains and empower the model to reason by maintaining semantic relationships. LGSR is composed of two parts, namely, local semantic reasoning (LSR) and global semantic reasoning (GSR). LSR utilizes weighted attribute similarity to develop accurate semantic relations among samples. Concretely, the feature is decomposed into several attributes, and the weights of the attributes are determined by the average intra-class similarity, thus suppressing the influence of class-irrelevant attributes. In contrast, GSR models the relation among the local prototypes of each class and facilitates the interaction of classes across domains. The prototypes aggregate semantic information from multiple samples, expanding the perceptual scope to the semantic space. 

Our contributions can be summarized as follows:

\begin{itemize}

    \item We delve into a practical and challenging topic, Single-DGOD, which is still relatively less investigated. Meanwhile, we propose a novel framework named SRCD to address two issues, one for eliminating the pseudo-correlation link between the single-source domain-specific attributes and category labels, and the other for maintaining the inherent semantic structural relationships among samples.
    
    \item SRCD consists of two key components, TBSA and LGSR. TBSA aims to eliminate the influence of irrelevant attributes such as texture, light, and shadow on category determination from the image level. The single-source domain is transformed into the compound domains by style transformation. LGSR uncovers the potential semantic structure among samples on instance-level features and activates the reasoning ability of the model by maintaining semantic relationships, thus enhancing cross-domain generalization.
    
    \item We evaluate the performance of our method under a variety of condition settings, including different weather, different cities, and virtual-to-reality situations. The experimental results demonstrate the effectiveness of the proposed SRCD.
    
\end{itemize}

\section{Related Work}

\subsection{Domain Generalization}
Domain generalization aims to train a robust enough model on several available source domains to generalize to unseen target domains~\cite{zhou2022domain}. The mainstream DG methods can be broadly classified into domain augmentation, distribution alignment and feature decoupling. Domain augmentation methods include image-wise augmentation~\cite{yue2019domain,zhou2020deep,zhou2020learning,xu2021fourier} and feature-wise augmentation~\cite{li2021simple,zhou2021mixstyle,li2022uncertainty}. For example, Zhou \textit{et al.}~\cite{zhou2020learning} leverage generative adversarial networks to generate samples with different styles, while Li \textit{et al.}~\cite{li2022uncertainty} synthesizes new forms of distributions based on the principle that feature statistics can characterize the style of an image. The core idea of distribution alignment is to drive the model to extract domain-invariant features. To achieve this goal, it is common practice to use metric learning~\cite{li2018domainMMD} or adversarial learning~\cite{li2018domain,matsuura2020domain} to bridge the feature differences across domains. For example, Li \textit{et al.}~\cite{li2018domainMMD} align different distributions by constraining the Maximum mean discrepancy (MMD) to be minimized. Matsuura \textit{et al.}~\cite{matsuura2020domain}, on the other hand, set up a domain classifier and reduce the distribution discrepancy by adversarial learning. Feature decoupling aims to decouple sample features into domain-invariant and domain-specific parts. For example, Khosla \textit{et al.}~\cite{khosla2012undoing} decompose the network parameters into domain-invariant and domain-specific lower-order components. Piratla \textit{et al.}~\cite{piratla2020efficient} develop a new method based on a low-public specific low-rank decomposition algorithm to adjust the final classification layer of the network. In addition, Carlucci \textit{et al.}~\cite{carlucci2019domain} propose a jigsaw game to improve the generalization of the model. Chen \textit{et al.}~\cite{chen2022compound} use the graph convolutional network to empower the inter-class inference ability of the model. Although all the above DG methods have achieved sound results, they are difficult to be directly applied to the areas of one single domain or object detection.

\subsection{Single-Domain Generalization}
Single-domain generalization considers the case where only one source domain is used for training. Most current methods~\cite{qiao2020learning,li2021progressive,fan2021adversarially} are customized for the classification task and cannot be easily transferred to the object detection task. For Single-DGOD, Wu \textit{et al.}~\cite{wu2022single} are the first to propose the Single-DGOD problem and provide a circular self-decoupling scheme. The scheme filters noisy information by secondary decoupling of domain-invariant features and, meanwhile, imposes consistency constraints on multilevel features. Zhao \textit{et al.}~\cite{zhao2022shade} propose a unified framework to learn generalized feature representations by means of dual consistency constraints on stylized and historical samples. Their methods focus on how to learn compact semantic representation, but ignore the attribute pseudo-correlation and the semantic structure. Our approach strives to learn invariant representation while maintaining intra- and inter-class semantic relationships, enabling the model to possess semantic reasoning capability.

\subsection{Multi-Domain Generalized Object Detection}
Lin \textit{et al.}~\cite{lin2021domain} are the first to explore the ability to enhance model generalization across domains in the object detection task. They present a novel feature decoupling framework that decomposes pixel-level and semantic-level features into domain-dependent and domain-independent parts. Zhang \textit{et al.}~\cite{zhang2022gated} propose gated decoupling networks, whose main idea is to adaptively turn on or off certain feature channels, thus focusing on the domain-invariant parts. Although their approaches have achieved promising results, they all rely on multiple domains as well as domain labels for support and are not applicable to the single-domain problem.

\begin{figure}
    \centering
    \begin{minipage}{\linewidth}
    \includegraphics[width=0.99\textwidth]{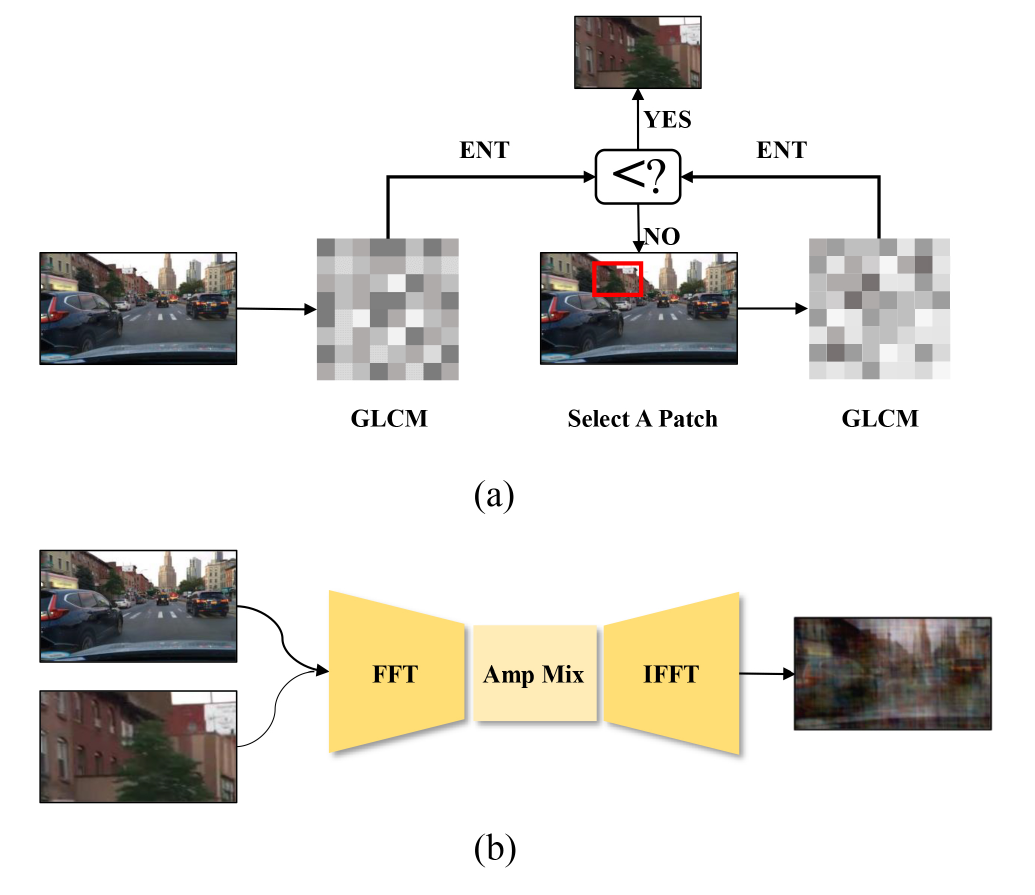}
    \end{minipage}
    \caption{Illustration of TBSA. (a) The procedure to select a patch, where \emph{ENT} denotes the entropy of the grey level co-occurrence matrix (GLCM). (b) The procedure of texture synthesis, where \emph{FFT} is the Fourier transformation, \emph{IFFT} is the inverse Fourier transformation, and \emph{Amp Mix} performs the mixup strategy on the amplitude spectrum.}
    \label{fig:xiaotu2}
\end{figure}

\section{Methodology}

\begin{figure*}
    \centering
    \begin{minipage}{\linewidth}
    \includegraphics[width=0.99\textwidth]{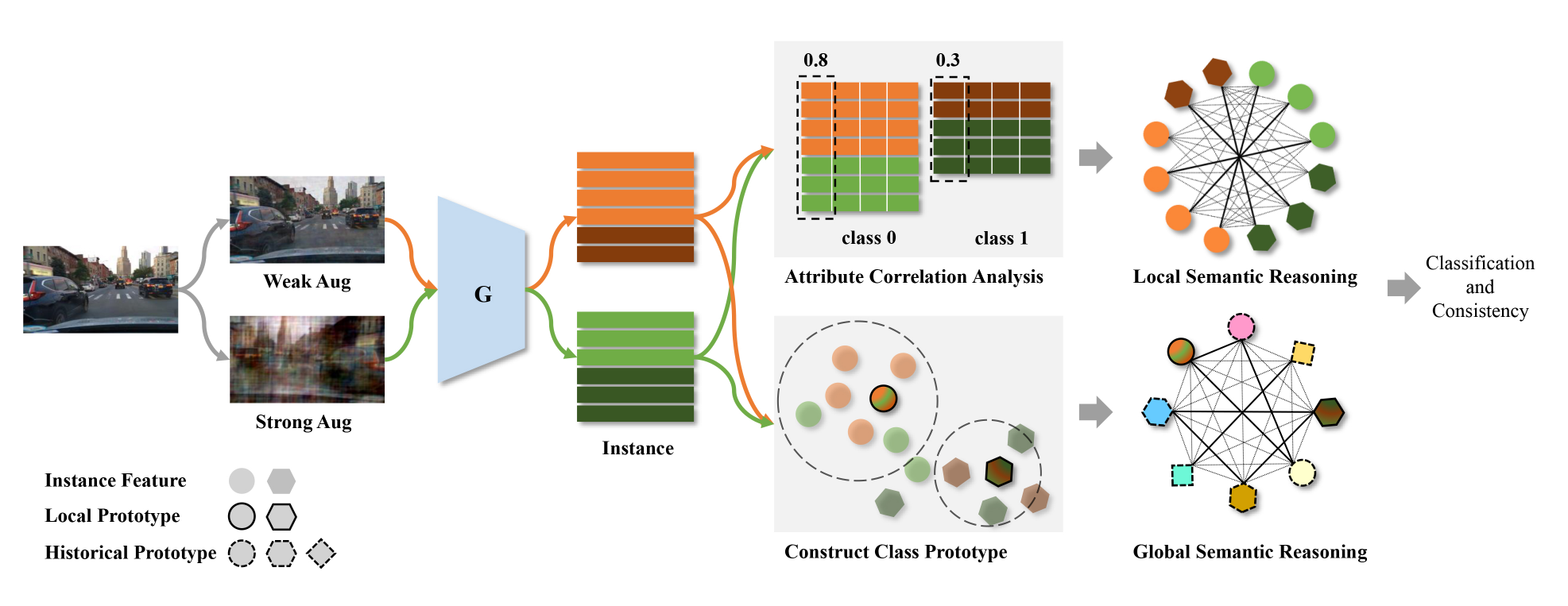}
    \end{minipage}
    \caption{The overall framework of SRCD. \emph{G} denotes the feature extractor. TBSA performs weak and strong augmentation on the input image. Then the feature extractor outputs their instance features, i.e., semantic-level features. LGSR models the semantic relation of the instance features to uncover the semantic structure.}
    \label{fig:framework}
\end{figure*}

The overall framework of the proposed method is shown in Fig.~\ref{fig:framework}. Our method is based on Faster-RCNN~\cite{faster2015towards} framework. Firstly, TBSA converts the input images into different styles by weak augmentation and strong augmentation, encouraging the model to utilize domain-invariant information for discrimination. Subsequently, the feature extraction network acquires the instance-level features of the images. LSR creates a refined relational graph of the samples within the batch. The correct semantic links between samples are obtained by attribute association analysis. GSR models global relationships among the current and historical category prototypes with a wider perceptual field. LGSR promotes the generalization ability of the model by uncovering and maintaining the semantic structure among samples in the compound domains.

\subsection{Texture-Based Self Augmentation}

Extracting domain-invariant information from a single source domain is critical and difficult. One of the challenges is the potential pseudo-correlation between source domain-specific attributes and labels. For example, vehicles in sunny days have shadows while cloudy and rainy days do not. TBSA aims to remove the pseudo-correlation between irrelevant attributes such as light and shadow, color, and labels at the image level. The phase spectrum of an image is known to carry more high-level semantic information, which is often not easily changeable. The magnitude spectrum, on the other hand, represents lower-order information such as texture~\cite{hansen2007structural,oppenheim1979phase,oppenheim1981importance}. Our goal is to encourage the model to learn semantic knowledge and ignore low-order information. To this end, TBSA performs pixel-level perturbation with the texture of image patches to force the model to focus on semantic content. The illustration of TBSA is shown in Fig.~\ref{fig:xiaotu2}.

We observe that the background of an image occupies more area in the object detection task. The elements in the background are complex and variable, and they possess rich textures, colors, etc. Therefore, the image itself is a texture library for augmentation. A simple approach is to randomly select a patch from an image and transpose its texture to the whole image. However, some patches are too plain, such as the sky and the road. We need to design a selection mechanism to evaluate the complexity of the texture to filter out the plain patches. For this purpose, we introduce the gray level co-occurrence matrix (GLCM)~\cite{haralick1973textural}, a traditional texture evaluation method. For a RGB image $X\in\mathbb{R}^{3\times H\times W}$, where $H, W$ indicate the height and width. Firstly convert $X$ to a grayscale image $\bar{X}\in\mathbb{R}^{H\times W}$, then its GLCM is represented as,
\begin{equation}
    G(i,j|d,\theta) = \sum_{a=0}^{H}\sum_{b=0}^{W}(\bar{X}_{a, b}=i)(\bar{X}_{a+d\cos\theta,b+d\sin\theta}=j),
\end{equation}
where $d$ denotes the relative distance and $\theta$ denotes the direction. Different GLCM can be obtained by setting $d$ and $\theta$. In this paper, we set $d=1$ and $\theta=0^o$. The GLCM describes the location distribution properties of pixels and its statistics are often used to quantify the texture characteristics of an image. The entropy of the GLCM measures the randomness and complexity of an image. The formula is,
\begin{equation}
    ENT = -\sum_{i=0}\sum_{j=0}G(i,j)\log(G(i,j)).
\end{equation}
We use $ENT$ to measure the complexity of the image to filter out plain patches. For one patch $P$ from the image $X$, if $ENT(P) < ENT(X)$, then discard this patch and reselect.

Once a suitable patch is selected, the picture is augmented by Fourier transformation. Specifically, the image $X$ and the patch $P$ are firstly Fourier transformed to obtain their respective amplitude spectra $Amp(X), Amp(P)$ and phase spectra $Pha(X), Pha(P)$. Then we perform the mixup strategy on their amplitude spectra and generate the augmented image by inverse Fourier transformation. The formula is expressed as,
\begin{gather}
    M_{amp} = (1-\phi)Amp(X)+\phi Amp(P), \\
    X_{aug} = \mathcal{F}^{-1}(M_{amp},Pha(X)),
\end{gather}
where $X_{aug}$ denotes the augmented image and $\mathcal{F}^{-1}$ is inverse Fourier transformation. $\phi$ is a random number. At each iteration, we perform two kinds of augmentation on the input image, weak augmentation, i.e. $\phi\in[0, 0.5)$ and strong augmentation i.e. $\phi\in[0.5, 1)$ and horizontal flip.

\subsection{Local-Global Semantic Reasoning}

Benefiting from TBSA, the single source domain is converted into the compound domains. An intuitive idea is to align the features from different domains to make the semantic space more compact, which is a common practice in many studies~\cite{dou2019domain, yao2022pcl, zhao2022shade, wu2022single}. However, such alignment approaches only emphasize intra-class distance but ignore the latent semantic structure. The reduced discrepancy between features may come from class-irrelevant attributes, which is caused by attribute pseudo-association. Meanwhile, the semantic association between classes is not taken into account. To this end, we propose LGSR to uncover the semantic relation among samples. We argue that maintaining such semantic relations while narrowing the gap between domains facilitates the model to perform semantic reasoning, thus improving cross-domain generalization. LGSR consists of two parts, Local Semantic Reasoning (LSR) and Global Semantic Reasoning (GSR).

\subsubsection{Local Semantic Reasoning}
LSR models the samples of the current batch. In each iteration, we feed an image $X$ to the network. The image is augmented by TBSA to obtain two samples $X_1, X_2$ with the same semantics but from different domains. The samples are fed into the convolutional network and then the instance (semantic) features are extracted by ROI-Pooling. We define their semantic features as $\mathbb{V}_1=\{v_1^1, v_2^1, ..., v_m^1|v\in\mathbb{R}^{C\times H\times W}\}, \mathbb{V}_2=\{v_1^2, v_2^2, ..., v_n^2|v\in\mathbb{R}^{C\times H\times W}\}$, where $m, n$ denotes the number of their instances. $C$ represents the number of feature channels. Here, $H$ and $W$ denote the height and width of the feature map.

Taking the instances in $\mathbb{V}_1,\mathbb{V}_2$ as nodes, our goal is to construct a relation graph across domains. The relation among the nodes is quantified by the cosine similarity of the features. However, as mentioned before, class-irrelevant attributes may mislead the relation construction. To this end, we decompose the features into several attributes and estimate the importance of each attribute separately. Specifically, for an instance feature $v\in\mathbb{R}^{C\times H\times W}\}$, it is flattened to $v\in\mathbb{R}^{CHW}\}$ and decomposed into $k$ segments. Then an instance feature is denoted as,
\begin{equation}
    v = [\bar{v}_1, \bar{v}_2, ..., \bar{v}_k],
\end{equation}
$\bar{v}\in\mathbb{R}^{CHW/k}$ stands for attribute. Then the distance between any two instance features $v_i, v_j$ is expressed as,
\begin{equation}
    S(v_i\in Q, v_j)=\frac{\sum_{g=1}^k\varepsilon_g^Q\cos(\bar{v}_g^i, \bar{v}_g^j)}{\sum_{g=1}^k\varepsilon_g^Q}.
\end{equation}
In the above equation, $S(\cdot)$ denotes the feature distance, $Q$ denotes the category label, that is, $S$ is related to the category $Q$ and $S(v_i, v_j)\neq S(v_j,v_i)$. $\varepsilon^Q=[\varepsilon_1^Q,...,\varepsilon_k^Q]$ denotes the weights of the attributes with respect to the category $Q$. Next, we describe how to calculate $\varepsilon^Q$. For any one attribute $\varepsilon_g^Q$, it is calculated by, 
\begin{equation}
    \varepsilon_g^Q = \frac{1}{|v\in Q|}\sum_{v_i\in Q,v_j\in Q}\cos(v_g^i,v_g^j).
\end{equation}
The reason for measuring the attribute weights in terms of average intra-class similarity is that, we argue that the average similarity will be smaller for complex and variable attributes such as background. However, the value of $\varepsilon^Q$ cannot be accurately estimated from just one batch of data, so we update it with an exponential moving average strategy,
\begin{equation}
    \varepsilon^Q_{(t)} = (1-\gamma)\varepsilon^Q_{(t-1)}+\gamma\varepsilon^Q_{(t)},
\end{equation}
where $t$ denotes the number of training iteration and $\gamma$ is the exponential decay rate and we set it to $0.99$. Then we instantiate the relation graph with an adjacency matrix,
\begin{equation}
    \mathcal{A^L}=\begin{bmatrix} \textbf{0}_{m\times m} & S(\mathbb{V}_1, \mathbb{V}_2)_{m\times n} \\ S(\mathbb{V}_2, \mathbb{V}_1)_{n\times m} & \textbf{0}_{n\times n} \end{bmatrix},
\end{equation}
where $\textbf{0}$ denotes the all-zero matrix. Finally, with the relation graph, we can perform information fusion to obtain new features with respect to the semantic structure,
\begin{equation}
    \mathbb{V}_{graph}=(\mathcal{A^L}+I)\bigotimes\mathbb{V},
\end{equation}
where $\mathbb{V}=[\mathbb{V}_1, \mathbb{V}_2]$, $\bigotimes$ denotes the matrix multiplication and $I$ denotes a diagonal matrix with diagonal elements of $1$ and the remaining elements of $0$.

The new and original features are fed into the downstream network together for classification. Their classification outputs are denoted as $O_{graph}^L$ and $O^L$, and we narrow the gap between them by minimizing the Kullback–Leibler (KL) divergence. The loss function is defined as,
\begin{equation}
    \mathcal{L}_{KL}^L=KL(O_{graph}^L||O^L).
\end{equation}
The classification loss of the new features is,
\begin{equation}
    \mathcal{L}_{CL}^L=CE(softmax(O_{graph}^L),y^L),
\end{equation}
where $CE$ is the cross-entropy loss. $y^L$ denotes the category label shared with $\mathbb{V}$. Then the total loss is defined as,
\begin{equation}
    \mathcal{L}_{LSR}=\mathcal{L}_{KL}^L+\mathcal{L}_{CL}^L.
\end{equation}

\subsubsection{Global Semantic Reasoning}
Limited by the batch size, LSR only models relationships for a small portion of samples and lacks the capability to perceive the global structure. To facilitate global perception and cross-domain interaction, GSR models the relation of local prototypes and historical prototypes. The prototype aggregates information from multiple samples. The class prototype of category $Q$ is defined as,
\begin{equation}
    \mathcal{P}^Q=\frac{1}{|v\in Q|}\sum_{v_i\in Q}v_i.
\end{equation}
Let $\mathcal{P}_1=\{\mathcal{P}_1^{Q_1},...,\mathcal{P}_1^{Q_r}\}$ denote the current prototype set and $r$ denotes the number of categories contained in the set. To broaden the perceptual field, we use a memory pool to cache the latest $Z$ prototype sets, i.e., historical prototypes. Let $\mathcal{P}_2, \mathcal{P}_3, ..., \mathcal{P}_{Z+1}$ denote historical prototype sets. 

Our goal is to construct a global relation graph with all prototypes of all prototype sets as nodes. Then the relation between any two prototypes is measured by the cosine similarity. However, as the training process advances forward, the cached historical prototypes may expire. Therefore, the pure cosine similarity cannot represent the relationship between the two well. To solve this problem, we weigh it by the storage time. Specifically, we use $T_1, T_2, T_3, \cdots, T_{z+1}$ to denote the length of time that each prototype set has been stored, i.e., the number of training iterations. Obviously, the value of $T_1$ is $0$. Then the distance between any two prototypes $\mathcal{P}_i, \mathcal{P}_j$ is expressed as,
\begin{equation}\label{eq:eq1}
    \bar{S}(\mathcal{P}_i, \mathcal{P}_j)=\exp^{-\frac{|T_i-T_j|}{\tau}}\cdot\cos(\mathcal{P}_i, \mathcal{P}_j),
\end{equation}
where $\tau$ is a temperature coefficient and we set it to $Z$. Let $\mathcal{\hat{P}}$ denote the super set containing all prototypes and the adjacency matrix $\mathcal{A^G}$ denote the global relation graph. $\mathcal{A^G}$ is computed by Eq. \ref{eq:eq1}. Then the structured prototype features are represented as,
\begin{equation}
    \mathcal{\hat{P}}_{graph}=\mathcal{A^G}\bigotimes\mathcal{\hat{P}}.
\end{equation}

The subsequent processing is the same as LSR. We naturally obtain the classification loss,
\begin{equation}
    \mathcal{L}_{CL}^G=CE(softmax(O^G),y^G)+CE(softmax(O_{graph}^G),y^G),
\end{equation}
and the KL loss,
\begin{equation}
    \mathcal{L}_{KL}^G=KL(O_{graph}^G||O^G).
\end{equation}
The meaning of the symbols is analogous to the LSR section and no more tautology here. Finally, the total loss is,
\begin{equation}
    \mathcal{L}_{GSR}=\mathcal{L}_{CL}^G+\mathcal{L}_{KL}^G.
\end{equation}

\subsection{Overall Objective}

Suppose the basic optimize objective of Faster-RCNN~\cite{faster2015towards} is defined as $\mathcal{L}_{det}$, including the classification and bounding box regression losses. The overall objective function is formulated as:
\begin{equation}
    \mathcal{L}_{SRCD}=\mathcal{L}_{det}+\lambda\mathcal{L}_{LSR}+\beta\mathcal{L}_{GSR},
\end{equation}
where $\lambda$ and $\beta$ are hyper-parameters.

\section{Experiments}

\subsection{Datasets}

To fully evaluate the effectiveness of our approach, we conducted experiments on multiple datasets, including different weather, different cities, and virtual-to-reality datasets. 

Different Weather: \textbf{Daytime-Sunny} contains 19,395 training images collected from the BDD100K dataset~\cite{yu2020bdd100k} under clear weather during the daytime. \textbf{Night-Sunny} contains 26,158 images from clear weather at night, which are also sampled from the BDD100K dataset. \textbf{Dusk-Rainy} and \textbf{Night-Rainy} are collected from rainy weather and include 3501 and 2494 images, respectively. Finally, \textbf{Daytime-Foggy} contains a total of 3775 images of foggy days, which are collected from the FoggyCityscapes~\cite{sakaridis2018semantic} and Adverse-Weather~\cite{hassaballah2020vehicle} datasets. All datasets share 7 categories. Following CDSD~\cite{wu2022single}, we use Daytime-Sunny as the source domain to train the model and then directly test on other domains. 

Different City: \textbf{Cityscapes}~\cite{cordts2016cityscapes} is collected from Germany and surrounding countries and contains 2975 images. \textbf{BDD100K}~\cite{yu2020bdd100k} is collected from the US and contains 100k images, of which we use 47,060 images from daytime clear weather. \textbf{KITTI}~\cite{geiger2012we} is also collected from Germany and contains 7,481 labeled images. Cityscapes and BDD100K share 7 categories. For KITTI, we only report the results of \emph{Car}. We use Cityscapes as the source domain and the other two as the target domains. 

Virtual-To-Reality: \textbf{Sim10K}~\cite{johnson2016driving} contains 10,000 virtual images rendered by the computer game Grand Theft Auto V (GTA V). We use Sim10K as the source domain, and Cityscapes, BDD100K, and KITTI as the target domains, and report the results of Car.

\subsection{Implementation Details}
Following CDSD~\cite{wu2022single}, we use the Faster-RCNN~\cite{faster2015towards} with the pre-trained Resnet-101~\cite{he2016deep} as backbone to experiment. The batch size is set to 1 and the shorter side of the input image is resized to 600. We optimize the network with stochastic gradient descent(SGD) with a momentum of 0.9 and the initial learning rate is set to 1e-3, which is decreased to 1e-4 after 5 epochs. The number of attributes $k$ is set to 4, and the size of the memory bank $Z$ is set to 10. The hyper-parameters $\lambda$ and $\beta$ are fixed to 0.1 and 0.01. We report the mean average precision (mAP) with an IoU threshold of 0.5. All experiments are implemented by the Pytorch framework and trained with a TITAN RTX GPU.

\begin{table*}[htbp]
    \centering
    \caption{Experimental results (\%) of Daytime-Sunny to other weather. SHADE does not report the results of categories.}
    \centering
    \resizebox{\textwidth}{!}{
    \begin{tabular}{c c c c c c c c >{\columncolor{lightgray}}c || c c c c c c c c >{\columncolor{lightgray}}c}
    \toprule
    \multicolumn{9}{c||}{\textbf{\large{Night-Sunny}}} & \multicolumn{9}{c}{\textbf{\large{Dusk-Rainy}}} \\
    \hline
     Method & bus & bike & car & motor & person & rider & truck & mAP & Method & bus & bike & car & motor & person & rider & truck & mAP\\
     \hline
     Faster-RCNN~\cite{faster2015towards} & 37.7 & 30.6 & 49.5 & 15.4 & 31.5 & 28.6 & 40.8 & 33.5 & Faster-RCNN~\cite{faster2015towards} & 36.8 & 15.8 & 50.1 & 12.8 & 18.9 & 12.4 & 39.5 & 26.6\\
     SW~\cite{pan2019switchable} & 38.7 & 29.2 & 49.8 & 16.6 & 31.5 & 28.0 & 40.2 & 33.4 & SW~\cite{pan2019switchable} & 35.2 & 16.7 & 50.1 & 10.4 & \textbf{20.1} & 13.0 & 38.8 & 26.3\\
     IBN-Net~\cite{pan2018two} & 37.8 & 27.3 & 49.6 & 15.1 & 29.2 & 27.1 & 38.9 & 32.1 & IBN-Net~\cite{pan2018two} & 37.0 & 14.8 & 50.3 & 11.4 & 17.3 & 13.3 & 38.4 & 26.1 \\
     IterNorm~\cite{huang2019iterative} & 38.5 & 23.5 & 38.9 & 15.8 & 26.6 & 25.9 & 38.1 & 29.6 & IterNorm~\cite{huang2019iterative} & 32.9 & 14.1 & 38.9 & 11.0 & 15.5 & 11.6 & 35.7 & 22.8 \\
     ISW~\cite{choi2021robustnet} & 38.5 & 28.5 & 49.6 & 15.4 & 31.9 & 27.5 & 41.3 & 33.2 & ISW~\cite{choi2021robustnet} & 34.7 & 16.0 & 50.0 & 11.1 & 17.8 & 12.6 & 38.8 &25.9 \\
     CDSD~\cite{wu2022single} & 40.6 & \textbf{35.1} & 50.7 & 19.7 & 34.7 & \textbf{32.1} & \textbf{43.4} & 36.6 & CDSD~\cite{wu2022single} & 37.1 & 19.6 & \textbf{50.9} & \textbf{13.4} & 19.7 & 16.3 & \textbf{40.7} & 28.2 \\
     SHADE~\cite{zhao2022shade} & - & - & - & - & - & - & - & 33.9 & SHADE~\cite{zhao2022shade} & - & - & - & - & - & - & - & \textbf{29.5} \\
     SRCD(ours) & \textbf{43.1} & 32.5 & \textbf{52.3} & \textbf{20.1} & \textbf{34.8} & 31.5 & 42.9 & \textbf{36.7} & SRCD(ours) & \textbf{39.5} & \textbf{21.4} & 50.6 & 11.9 & \textbf{20.1} & \textbf{17.6} & 40.5 & 28.8 \\
     \hline
      \multicolumn{9}{c||}{\textbf{\large{Night-Rainy}}} & \multicolumn{9}{c}{\textbf{\large{Daytime-Foggy}}} \\
    \hline
     Method & bus & bike & car & motor & person & rider & truck & mAP & Method & bus & bike & car & motor & person & rider & truck & mAP\\
     \hline
     Faster-RCNN~\cite{faster2015towards} & 22.6 & 11.5 & 27.7 & 0.4 & 10.0 & 10.5 & 19.0 & 14.5 & Faster-RCNN~\cite{faster2015towards} & 30.7 & 26.7 & 49.7 & 26.2 & 30.9 & 35.5 & 23.2 & 31.9 \\
     SW~\cite{pan2019switchable} & 22.3 & 7.8 & 27.6 & 0.2 & 10.3 & 10.0 & 17.7 & 13.7 & SW~\cite{pan2019switchable} & 30.6 & 26.2 & 44.6 & 25.1 & 30.7 & 34.6 & 23.6 & 30.8 \\
     IBN-Net~\cite{pan2018two} & 24.6 & 10.0 & 28.4 & 0.9 & 8.3 & 9.8 & 18.1 & 14.3 & IBN-Net~\cite{pan2018two} & 29.9 & 26.1 & 44.5 & 24.4 & 26.2 & 33.5 & 22.4 & 29.6 \\
     IterNorm~\cite{huang2019iterative} & 21.4 & 6.7 & 22.0 & 0.9 & 9.1 & 10.6 & 17.6 & 12.6 & IterNorm~\cite{huang2019iterative} &29.7 & 21.8 & 42.4 & 24.4 & 26.0 & 33.3 & 21.6 & 28.4 \\
     ISW~\cite{choi2021robustnet} & 22.5 & 11.4 & 26.9 & 0.4 & 9.9 & 9.8 & 17.5 & 14.1 & ISW~\cite{choi2021robustnet} & 29.5 & 26.4 & 49.2 & 27.9 & 30.7 & 34.8 & 24.0 & 31.8 \\
     CDSD~\cite{wu2022single} & 24.4 & 11.6 & 29.5 & \textbf{9.8} & \textbf{10.5} & 11.4 & 19.2 & 16.6 & CDSD~\cite{wu2022single} & 32.9 & 28.0 & 48.8 & 29.8 & 32.5 & 38.2 & 24.1 & 33.5 \\
     SHADE~\cite{zhao2022shade} & - & - & - & - & - & - & - & 16.8 & SHADE~\cite{zhao2022shade} & - & - & - & - & - & - & - & 33.4 \\
     SRCD(ours) & \textbf{26.5} & \textbf{12.9} & \textbf{32.4} & 0.8 & 10.2 & \textbf{12.5} & \textbf{24.0} & \textbf{17.0} & SRCD(ours) & \textbf{36.4} & \textbf{30.1} & \textbf{52.4} & \textbf{31.3} & \textbf{33.4} & \textbf{40.1} & \textbf{27.7} & \textbf{35.9}  \\
     \bottomrule
    \end{tabular}}
    \label{tab:weather}
\end{table*}

\begin{table}[htbp]
    \centering
    \caption{Experimental results (\%) of Cityscapes to BDD100K.}
     \resizebox{0.48\textwidth}{!}{
    \begin{tabular}{c c c c c c c c >{\columncolor{lightgray}}c}
    \toprule
     Method & bike & bus & car & truck & rider & person & motor & mAP \\
     \hline
     Faster-RCNN~\cite{faster2015towards}  & 22.7 & \textbf{21.9} & 36.9 & 22.6 & 25.4 & 24.1 & 17.8 & 24.5 \\
     SW~\cite{pan2019switchable} & 22.2 & 21.3 & 36.7 & 20.9 & 25.6 & 23.1 & 18.8 & 24.1 \\
     IBN-Net~\cite{pan2018two} & 19.2 & 14.9 & 31.9 & 13.7 & 21.4 & 19.3 & 13.4 & 19.1 \\
     IterNorm~\cite{huang2019iterative} & 21.6 & 21.2 & 36.2 & 20.9 & 23.4 & 10.6 & 18.0 & 23.7 \\
     ISW~\cite{choi2021robustnet} & 21.6 & 20.9 & 35.2 & 17.8 & 22.7 & 22.4 & 16.6 & 22.5 \\
     CDSD~\cite{wu2022single} & 22.9 & 20.5 & 33.8 & 18.2 & 23.6 & 18.5 & 14.7 & 21.7 \\
     SHADE~\cite{zhao2022shade} & \textbf{25.1} & 19.0 & 36.8 & 19.8 & 24.9 & 24.1 & 18.4 & 24.0 \\
     SRCD(ours)  & 24.8 & 21.5 & \textbf{38.7} & \textbf{23.1} & \textbf{28.4} & \textbf{25.7} & \textbf{19.0} & \textbf{25.9}\\
    \bottomrule
    \end{tabular}}
    \label{tab:city-bdd}
\end{table}

\begin{table}[htbp]
    \caption{Experimental results (\%) of Cityscapes to KITTI.}
    \centering
    \begin{tabular}{c c}
    \toprule
     Method & AP of Car \\
     \hline
     Faster-RCNN~\cite{faster2015towards}  & 72.5 \\
     SW~\cite{pan2019switchable} & 72.9 \\
     IBN-Net~\cite{pan2018two} & 66.7 \\
     IterNorm~\cite{huang2019iterative} & 71.9 \\
     ISW~\cite{choi2021robustnet} & 71.4 \\
     CDSD~\cite{wu2022single} & 70.5 \\
     SHADE~\cite{zhao2022shade} & 72.2 \\
     SRCD(ours)  & \textbf{73.2} \\
    \bottomrule
    \end{tabular}
    \label{tab:city-kitti}
\end{table}

\subsection{Main Results}

We compare with the following methods. SW~\cite{pan2019switchable}, IBN-Net~\cite{pan2018two}, IterNorm~\cite{huang2019iterative}, ISW~\cite{choi2021robustnet} improve the generalization of the model by designing different feature regularization methods. CDSD~\cite{wu2022single} uses cyclic self-decoupling to extract domain-invariant features. SHADE~\cite{zhao2022shade} learns robust feature representations by means of dual consistency constraints. We use their released code for experiments.

\noindent\textbf{Different Weather.} Table~\ref{tab:weather} shows the experimental results under different weather conditions. Daytime-Sunny is used as the source domain and other datasets are used as the target domains. It can be seen that there is a huge domain shift between clear and severe weather. Our method achieves the best results on three of the datasets. In particular, on the Daytime-Foggy dataset, our method outperforms Faster-RCNN by 4\% and outperforms CDSD by 2.4\%. Compared to the current state-of-the-art methods CDSD and SHADE, our method improves by 2.4\% and 2.5\%, respectively. The experimental results indicate that obtaining pure domain-invariant representations from the single-source domain is extremely challenging. And our method effectively mitigates the interference of irrelevant attributes on the discrimination, while the semantic relationship modeling has a beneficial effect on model transferability.

\noindent\textbf{Different City.} Due to the different styles of architecture, roads, etc., there is a domain shift problem between different cities. We conducted cross-domain experiments accordingly. Table~\ref{tab:city-bdd} shows the results of Cityscapes to BDD100K, and it can be seen that our method achieves the best results, getting a 3\% gain on the category of \emph{rider} and an overall gain of 1.4\%. The CDSD based on feature decoupling does not bring significant gain effect, which may be owing to the fact that the model misclassifies the source domain-specific features as domain-invariant features, indicating that a single-minded pursuit of invariant features may be a suboptimal choice. Table~\ref{tab:city-kitti} shows the experimental results of Cityscapes to KITTI. Since both datasets are from the same city, Faster-RCNN provides a strong baseline. Compared to other methods, our method still provides a small performance improvement.

\begin{table}[htbp]
    \caption{Experimental results (\%) of Sim10K to Real-World Datasets. Only results of \emph{Car} are reported.}
    \centering
    \begin{tabular}{c  c  c  c}
    \toprule
     Method & Cityscpaes & BDD100K & KITTI \\
     \hline
     Faster-RCNN~\cite{faster2015towards}  & 34.3 & 29.8 & 47.0 \\
     SW~\cite{pan2019switchable} & 34.5 & 30.0 & 47.2\\
     IBN-Net~\cite{pan2018two} & 33.2 & 25.7 & 48.1\\
     IterNorm~\cite{huang2019iterative} & 34.3 & 30.3 & 46.9\\
     ISW~\cite{choi2021robustnet} & 40.4 & 28.5 & 55.0\\
     CDSD~\cite{wu2022single} & 35.2 & 27.4 & 47.8\\
     SHADE~\cite{zhao2022shade} & 40.9 & 30.3 & 55.6\\
     SRCD(ours)  & \textbf{43.0} & \textbf{31.6} & \textbf{60.4}\\
    \bottomrule
    \end{tabular}
    \label{tab:sim10k-other}
\end{table}

\begin{table}[htbp]
    \caption{Ablation analysis of SRCD. N-S, D-R, N-R, D-F are Night-Sunny, Dusk-Rainy, Night-Rainy, Daytime-Foggy respectively.}
    \centering
    \resizebox{0.48\textwidth}{!}{
    \begin{tabular}{c c  c  c  c  c}
    \toprule
     Method & N-S & D-R & N-R & D-F & Avg. \\
     \hline
     Faster-RCNN~\cite{faster2015towards}  & 33.5 & 26.6 & 14.5 & 31.9 & 26.6\\
     \hline
     SRCD w/o TBSA & 35.3 & 27.4 & 15.8 & 33.4 & 28.0\\
     TBSA w/o GLCM & 34.8 & 28.1 & 16.2 & 34.7 & 28.5\\
     SRCD w/o LSR & 35.2 & 27.7 & 16.5 & 34.2 & 28.4\\
     SRCD w/o GSR & 36.2 & 27.6 & 15.7 & 35.1 & 28.7\\
     LSR w/o Attribute Analysis & 35.6 & 27.8 & 16.7 & 34.8 & 28.7\\
     GSR w/o Historical Prototypes & 35.9 & 28.4 & 16.5 & 35.0 & 29.0\\
     \hline
     SRCD  & \textbf{36.7} & \textbf{28.8} & \textbf{17.0} & \textbf{35.9} & \textbf{29.6}\\
    \bottomrule
    \end{tabular}}
    \label{tab:ablation-study}
\end{table}

\noindent\textbf{Virtual-To-Reality.} Inherent differences in distribution exist between synthetic and real data. To investigate the generalizability of the model on synthetic data, we conducted the corresponding experiments, and the results are shown in Table~\ref{tab:sim10k-other}. Our method leads the other methods by a large margin. In particular, on the KITTI dataset, the proposed SRCD outperforms Faster-RCNN by 13.4\% and SHADE by 4.8\%. On the Cityscapes dataset, our method is higher than Faster-RCNN by 8.7\%, demonstrating that our method effectively models the potential target domains and learns robust features through relational modeling.

\subsection{Further Empirical Analysis}

\begin{figure*}
    \centering
    
    \begin{minipage}{0.99\textwidth}
    \centering
    
    \begin{minipage}{0.24\textwidth}
    \includegraphics[width=44mm,height=26mm]{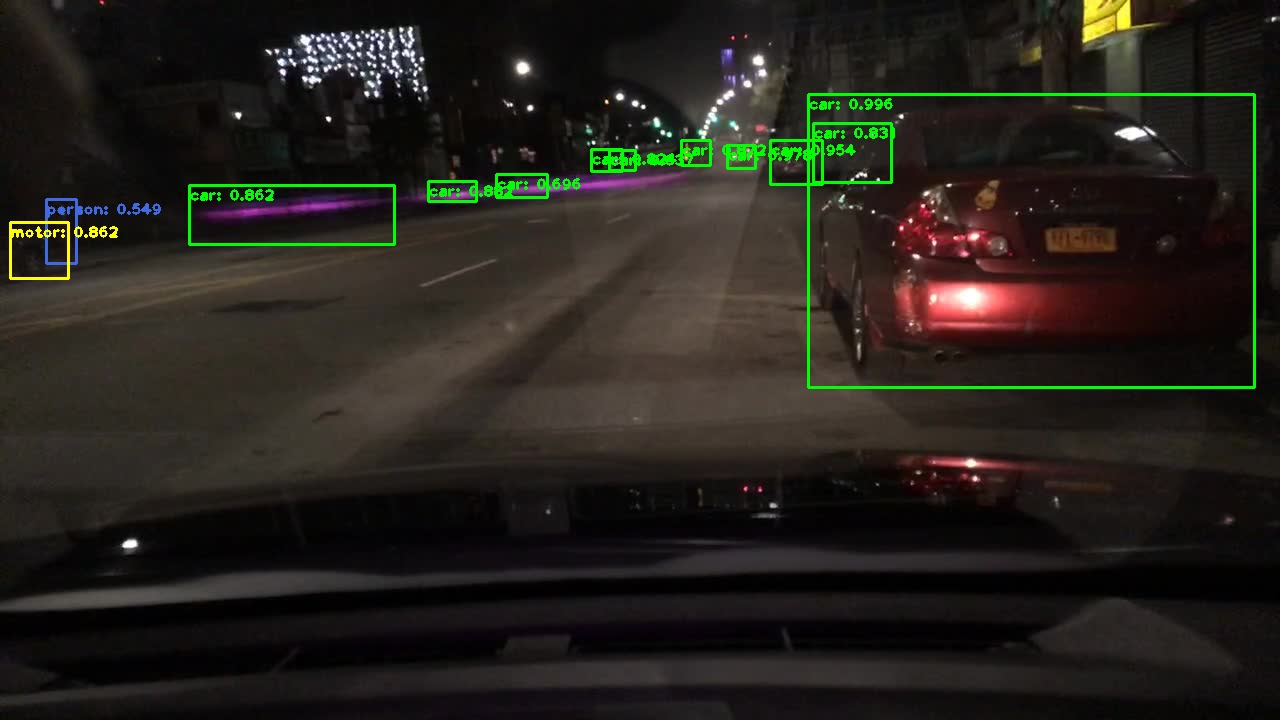}
    \end{minipage}
    \begin{minipage}{0.24\textwidth}
    \includegraphics[width=44mm,height=26mm]{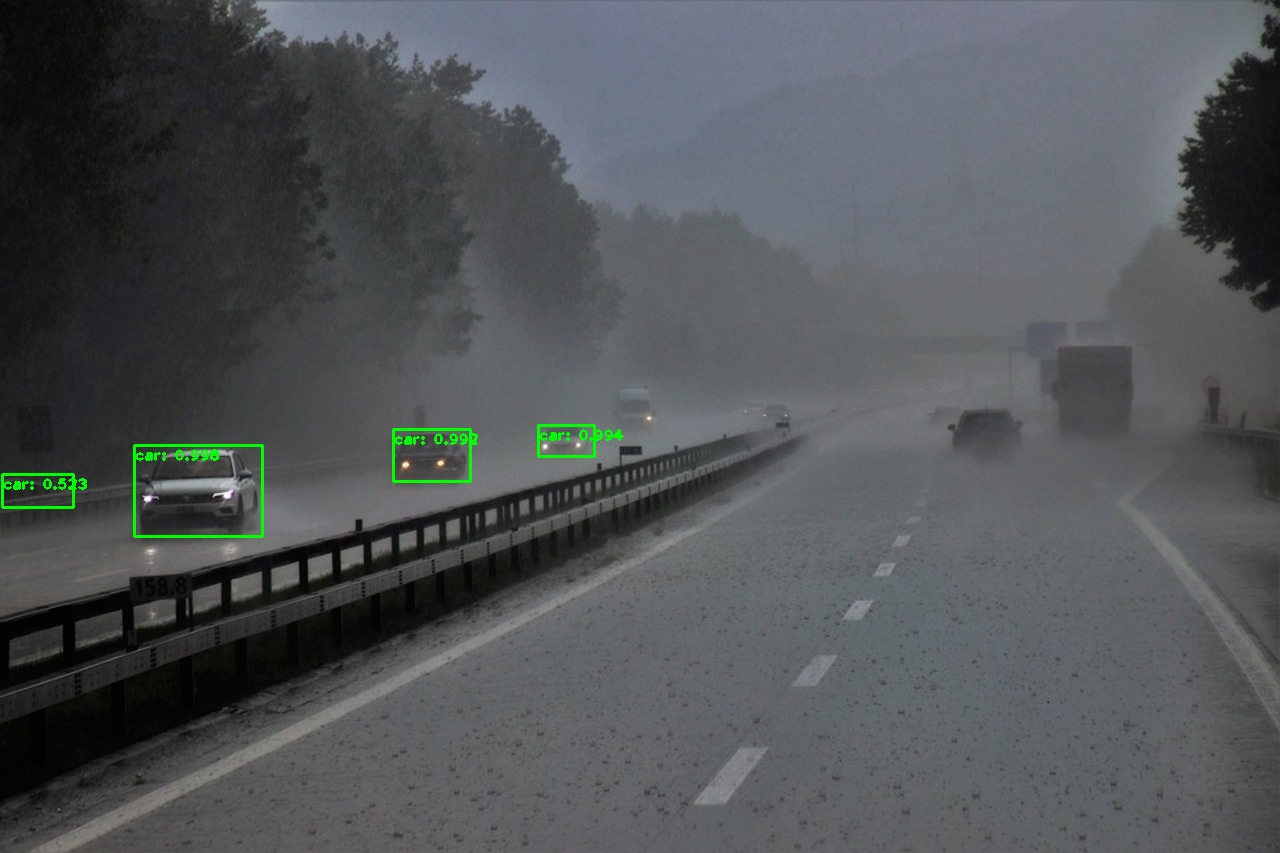}
    \end{minipage}
    \begin{minipage}{0.24\textwidth}
    \includegraphics[width=44mm,height=26mm]{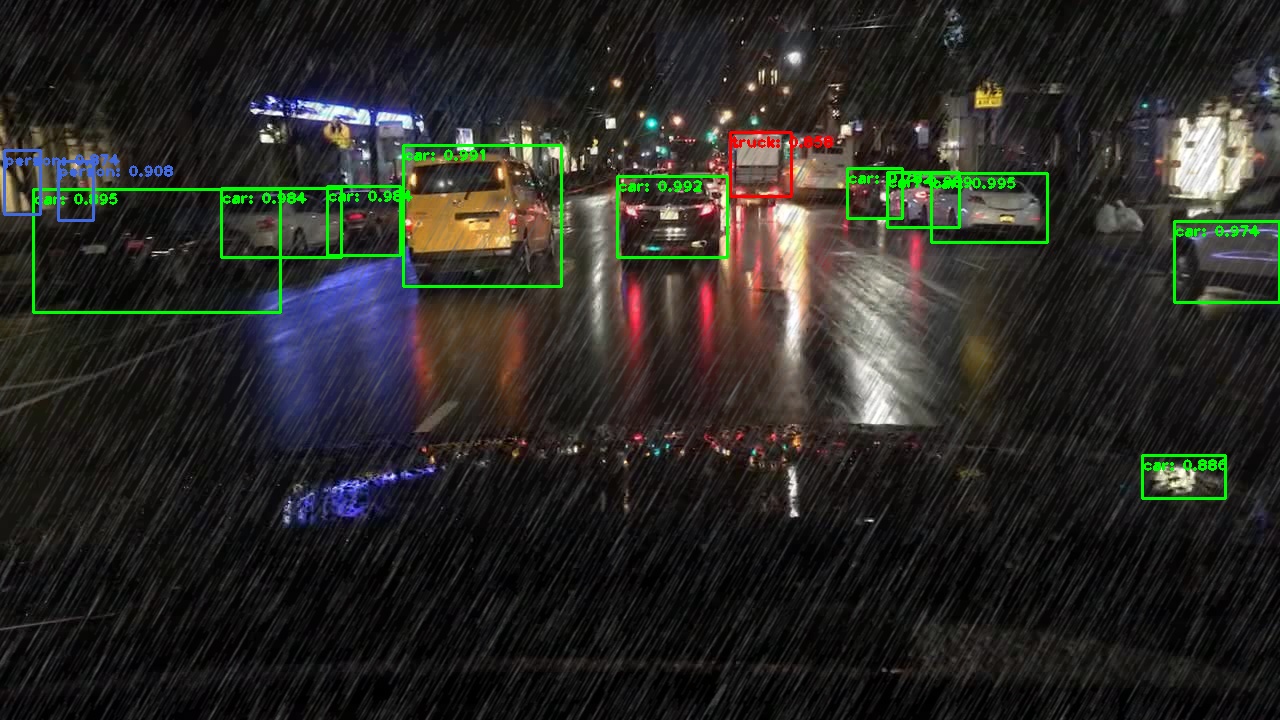}
    \end{minipage}
    \begin{minipage}{0.24\textwidth}
    \includegraphics[width=44mm,height=26mm]{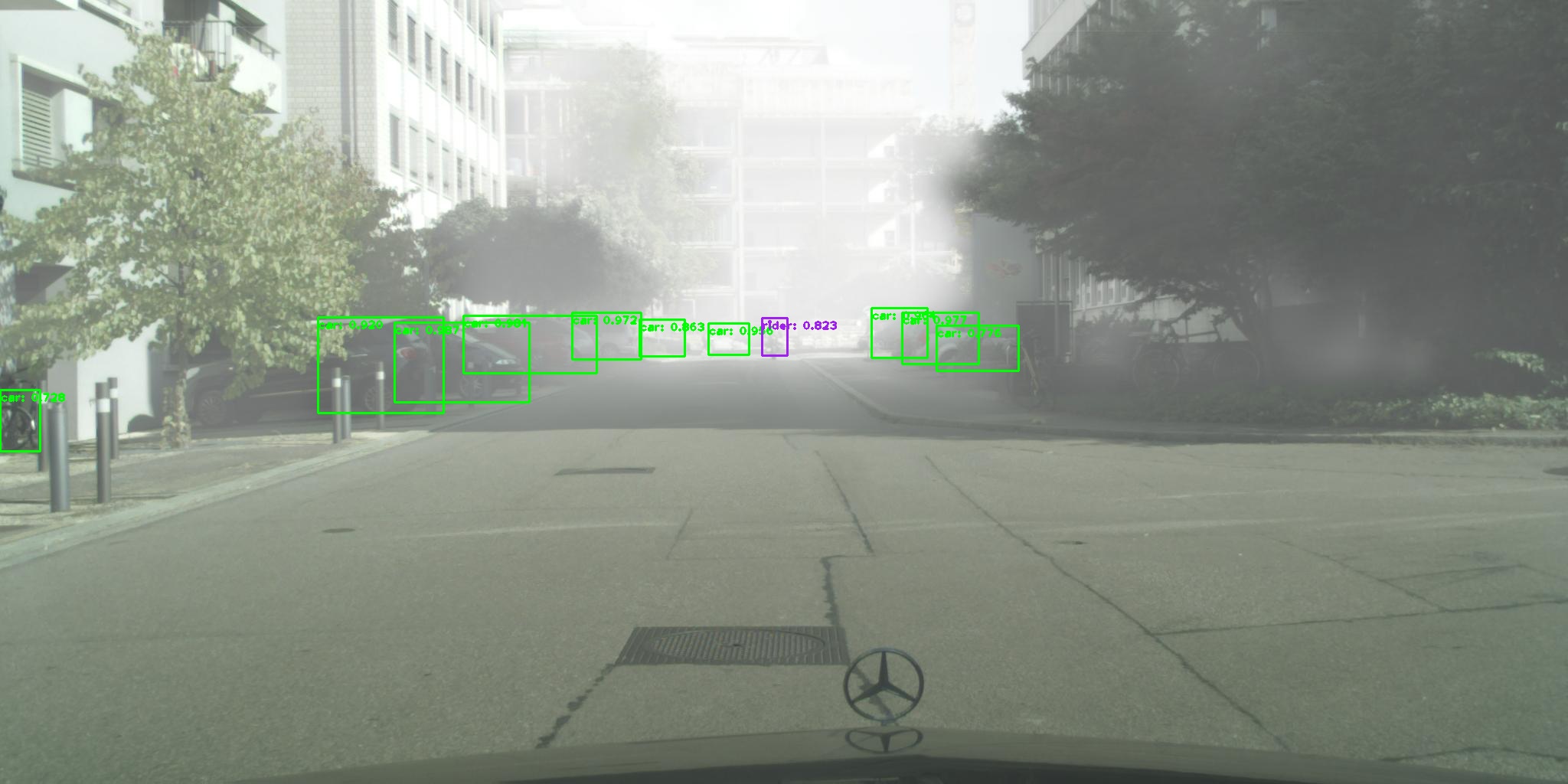}
    \end{minipage}

    \vspace{0.5mm}
    \begin{minipage}{0.24\textwidth}
    \includegraphics[width=44mm,height=26mm]{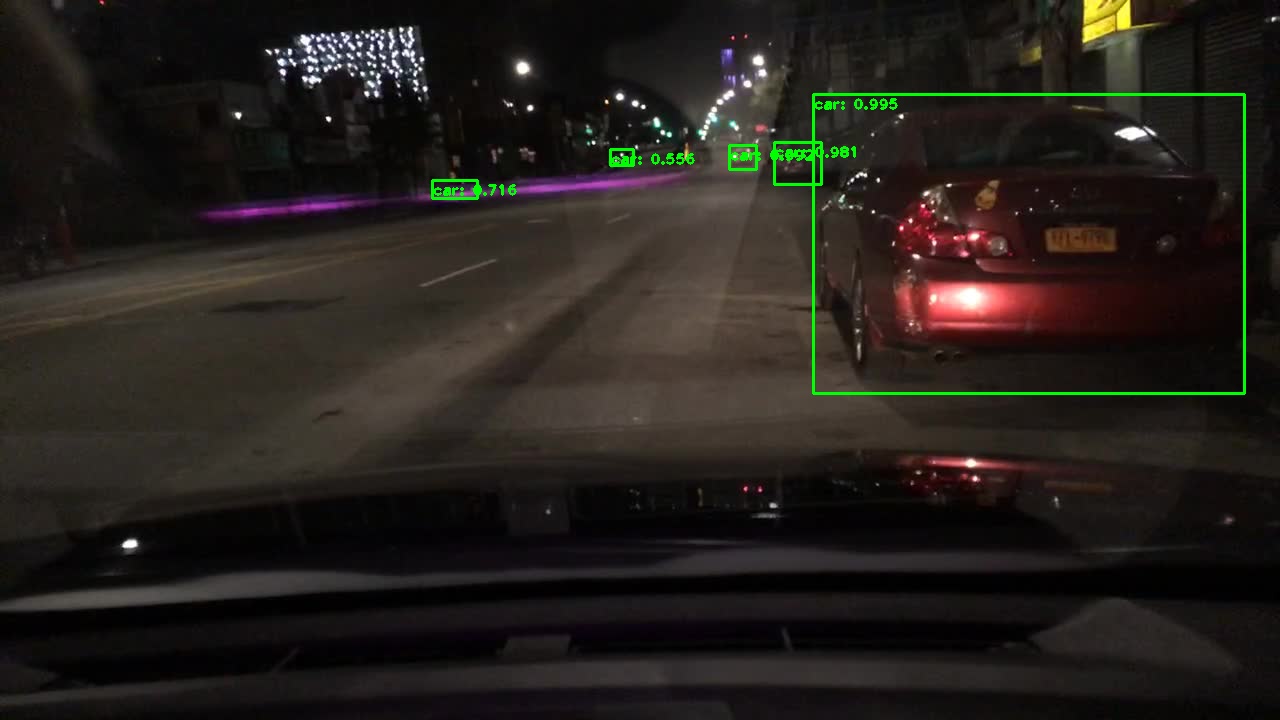}
    \end{minipage}
    \begin{minipage}{0.24\textwidth}
    \includegraphics[width=44mm,height=26mm]{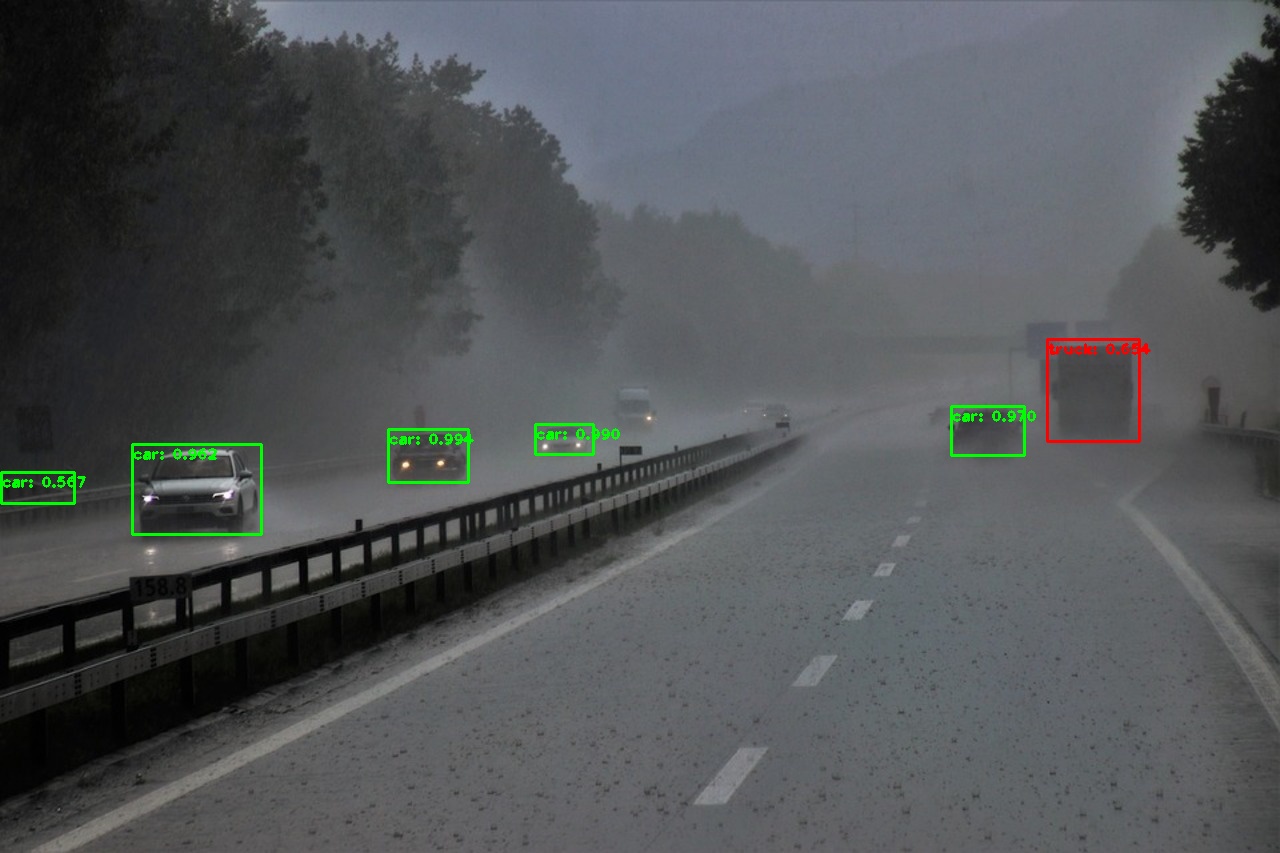}
    \end{minipage}
    \begin{minipage}{0.24\textwidth}
    \includegraphics[width=44mm,height=26mm]{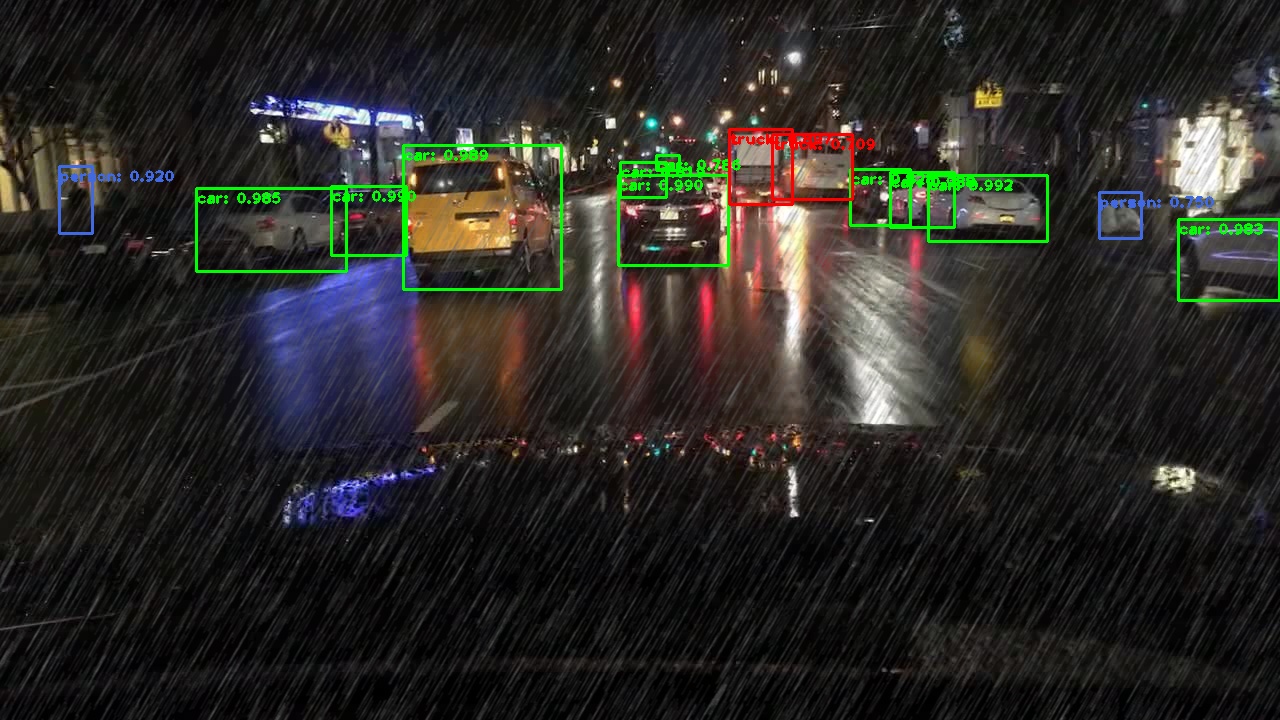}
    \end{minipage}
    \begin{minipage}{0.24\textwidth}
    \includegraphics[width=44mm,height=26mm]{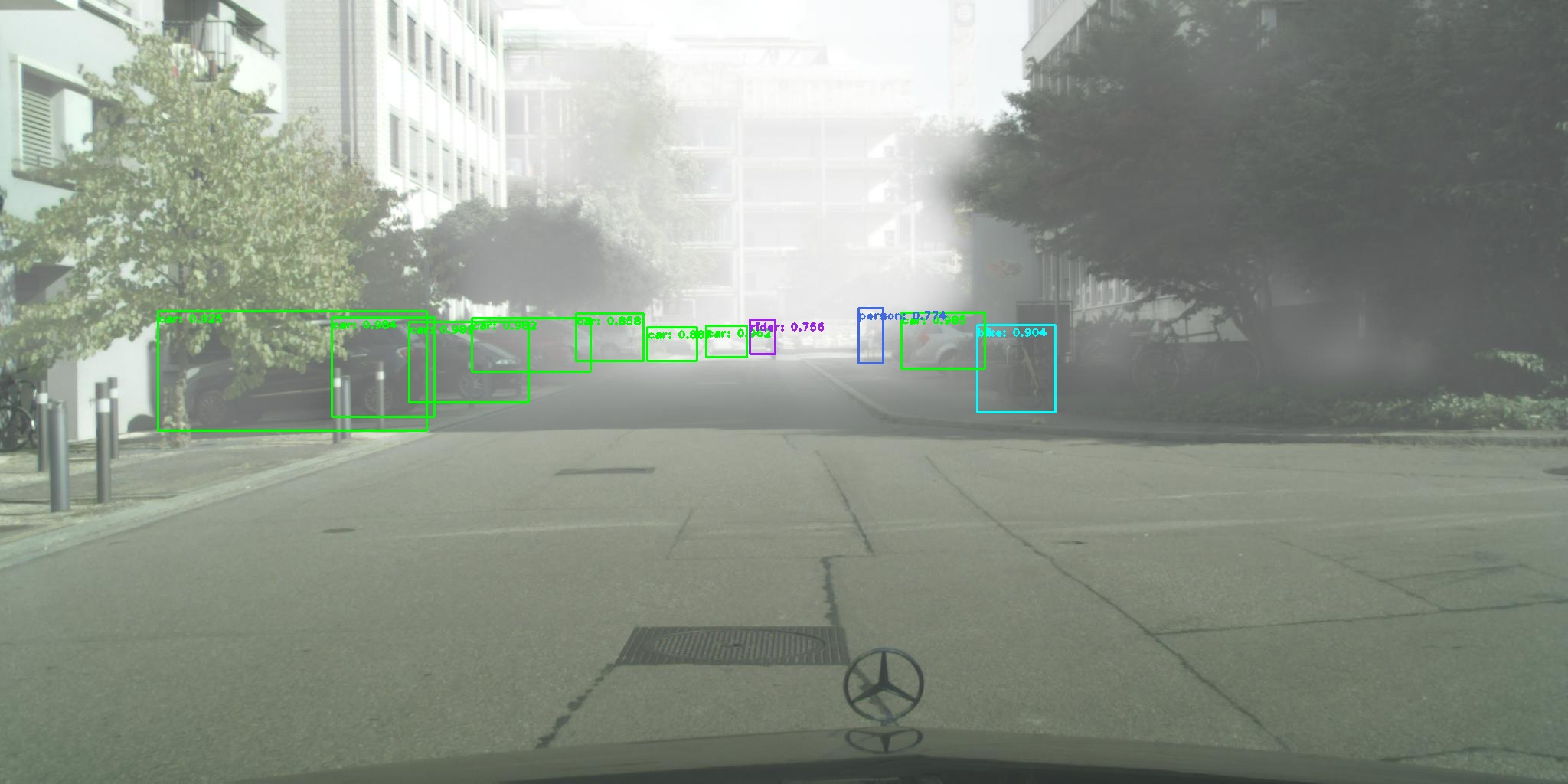}
    \end{minipage}

    \vspace{0.5mm}
    \begin{minipage}{0.24\textwidth}
    \includegraphics[width=44mm,height=26mm]{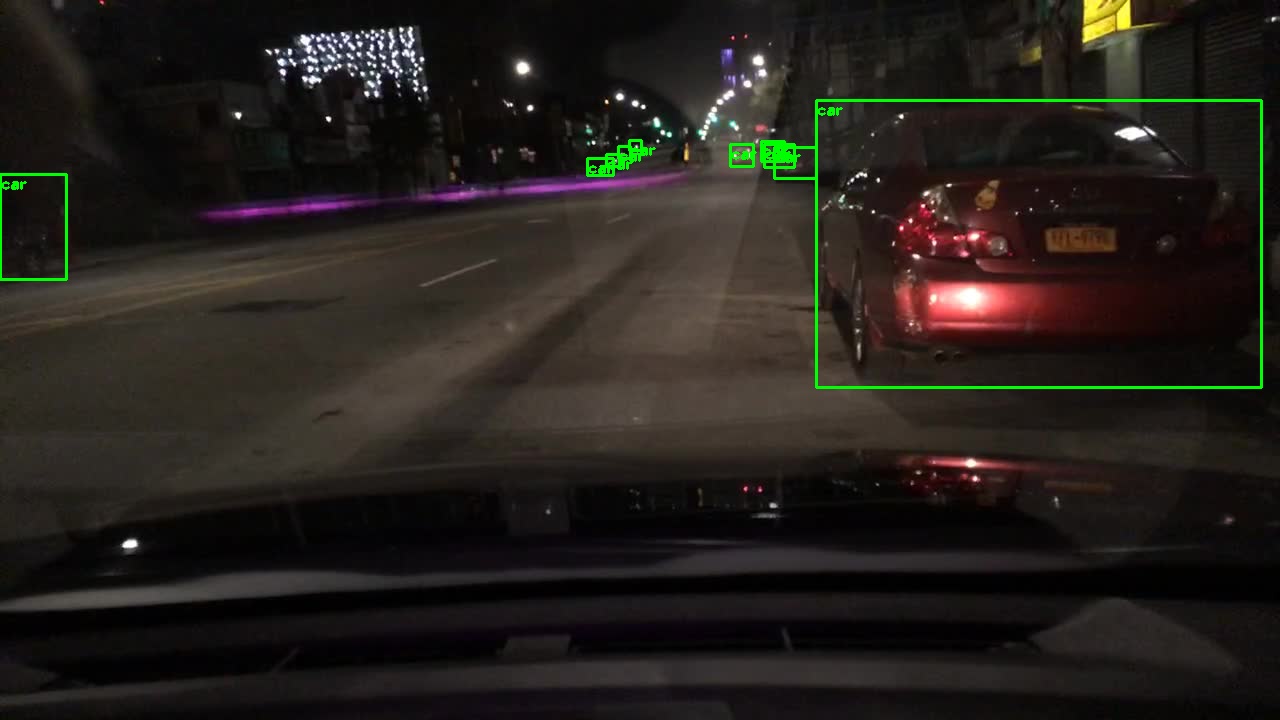}
    
    \centering
    \end{minipage}
    \begin{minipage}{0.24\textwidth}
    \includegraphics[width=44mm,height=26mm]{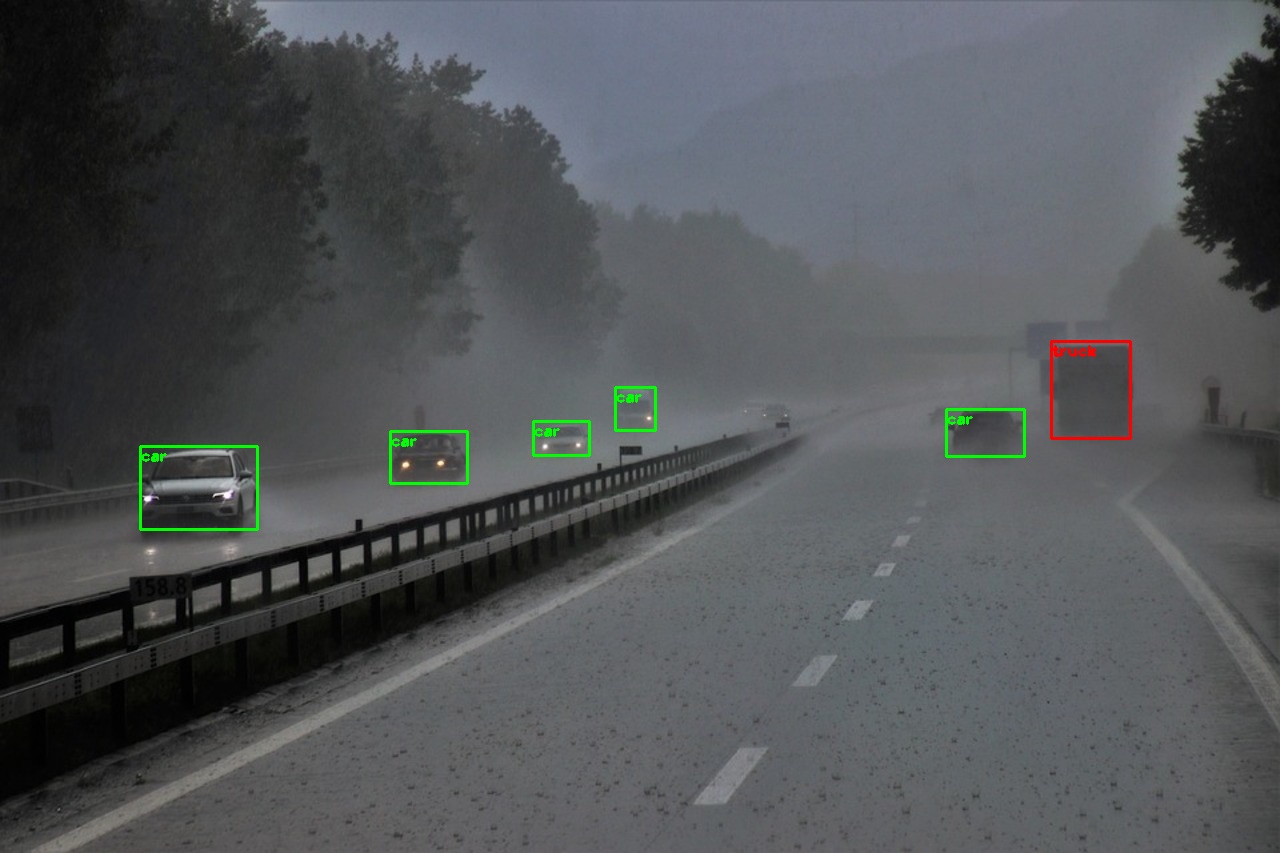}
    
    \centering
    \end{minipage}
    \begin{minipage}{0.24\textwidth}
    \includegraphics[width=44mm,height=26mm]{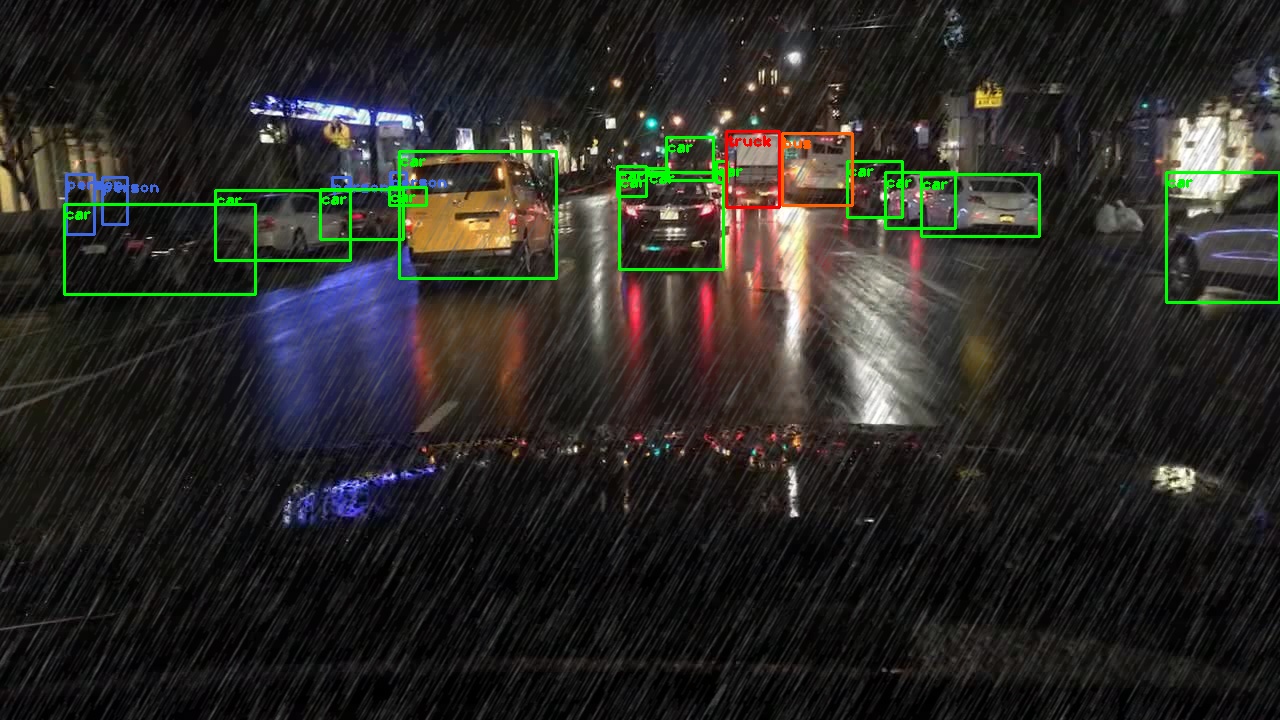}
    
    \centering
    \end{minipage}
    \begin{minipage}{0.24\textwidth}
    \includegraphics[width=44mm,height=26mm]{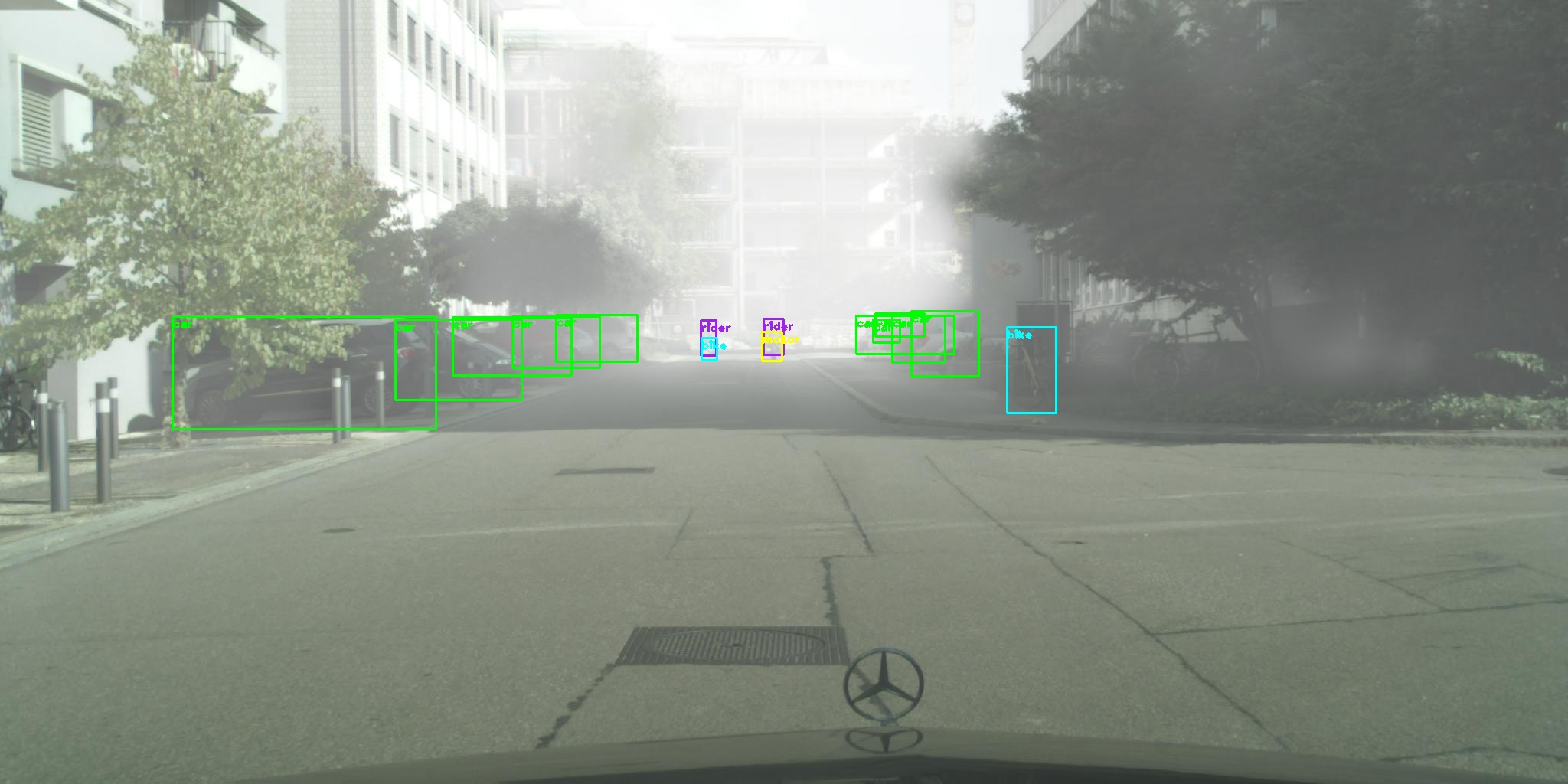}
    
    \centering
    \end{minipage}
    
    \caption{Qualitative detection results on Night-Sunny, Dusk-Rainy, Night-Rainy, Daytime-Foggy. \textbf{First Row:} The results of Faster-RCNN. \textbf{Second Row:} The results of our approach. \textbf{Third Row:} Ground truth.}
    \label{fig:visbox}
    \vspace{5mm}
    \end{minipage}
\end{figure*}

\noindent\textbf{Ablation study.} To further investigate the effectiveness of the individual components, a series of ablation experiments are performed. The experimental results are shown in Table~\ref{tab:ablation-study}, where w/o indicates the removal of the component. It can be seen that removing any one of the components degrades the performance, demonstrating that the design of each component is reasonable.

\begin{table}[htbp]
    \caption{Experimental results (\%) of Sim10K to Cityscapes. 'target' indicates whether the target domain is accessed during the training phase or not (\ding{51} is yes and \ding{55} is no).}
    \centering
    \begin{tabular}{c |c| c}
    \toprule
     Method & target & AP of Car \\
     \hline
     DAF~\cite{chen2018domain}  & \ding{51} & 38.9 \\
     SWDA~\cite{saito2019strong} & \ding{51} & 40.1 \\
     MAF~\cite{he2019multi} & \ding{51} & 41.1 \\
     HTCN~\cite{chen2020harmonizing} & \ding{51} & 42.5 \\
     DBGL~\cite{chen2021dual} & \ding{51} & 42.7 \\
     SRCD(ours)  & \ding{55} &  \textbf{43.0} \\
    \bottomrule
    \end{tabular}
    \label{tab:da}
\end{table}

\noindent\textbf{Comparison with domain adaptation methods.} To further explore the generalizability of the model, we conducted comparative experiments on Sim10K-to-Cityscapes with domain adaptation methods, which require access to the target domain during the training phase. The experimental results are shown in Table~\ref{tab:da}, and we can see that our method is ahead of them, demonstrating the strong cross-domain generalization ability of our method. It also shows that the generalization of the model has not been fully exploited.

\begin{figure}
    \centering
    
    \begin{minipage}{0.99\linewidth}
    \centering
    
    \begin{minipage}{0.24\linewidth}
    \includegraphics[width=22mm,height=12mm]{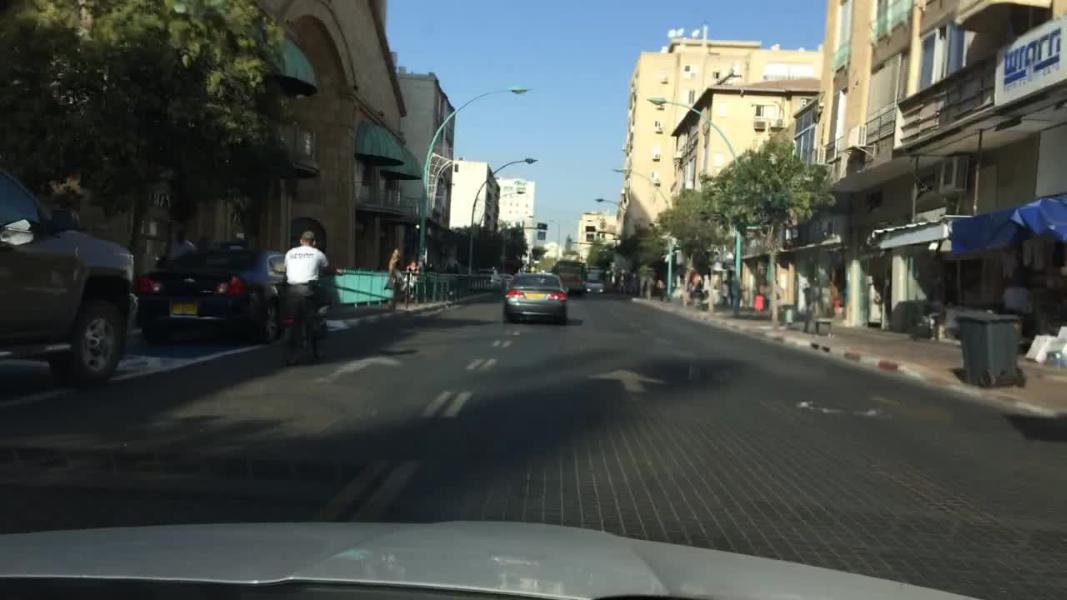}
    \end{minipage}
    \begin{minipage}{0.24\linewidth}
    \includegraphics[width=22mm,height=12mm]{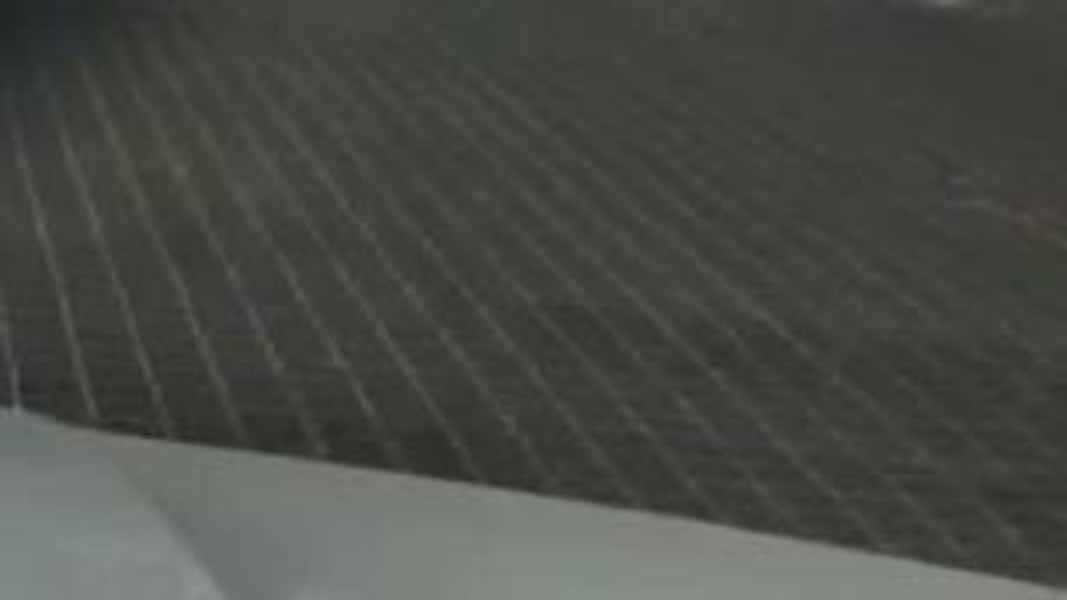}
    \end{minipage}
    \begin{minipage}{0.24\linewidth}
    \includegraphics[width=22mm,height=12mm]{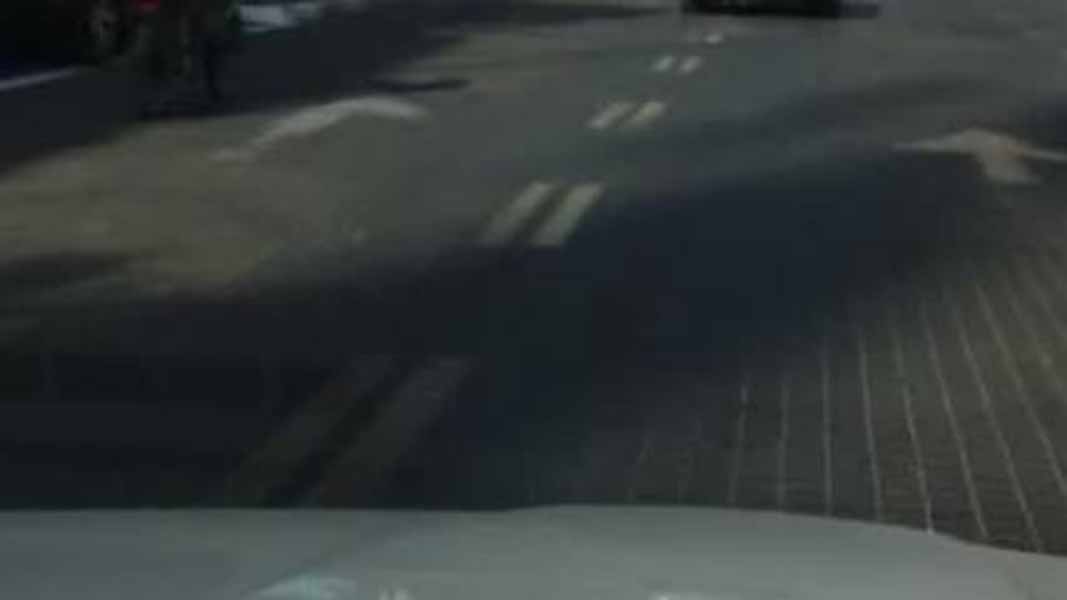}
    \end{minipage}
    \begin{minipage}{0.24\linewidth}
    \includegraphics[width=22mm,height=12mm]{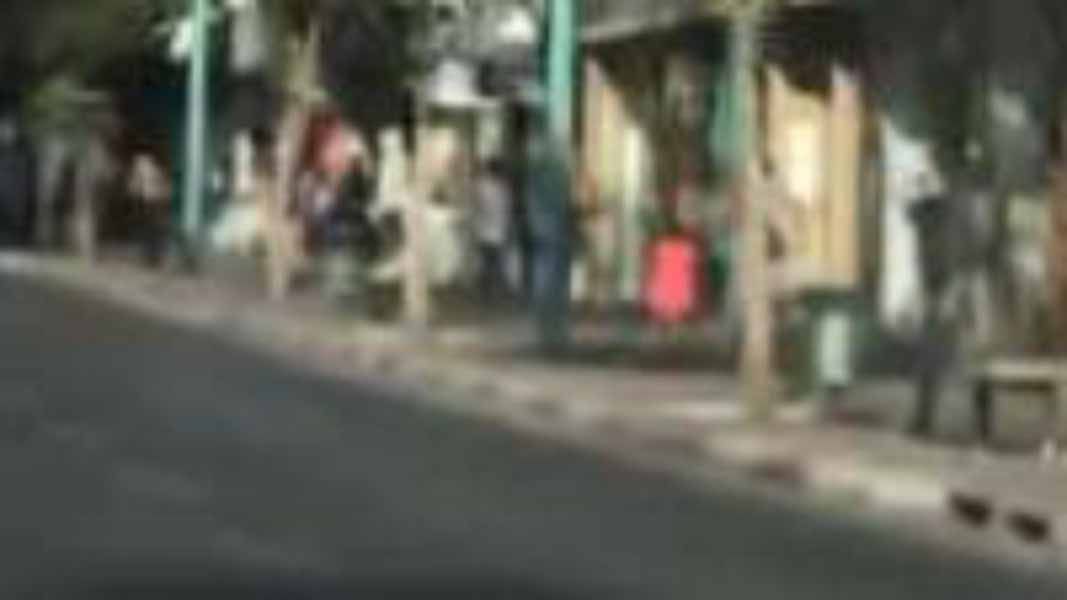}
    \end{minipage}

    \vspace{0.3mm}
    \begin{minipage}{0.24\linewidth}
    \includegraphics[width=22mm,height=12mm]{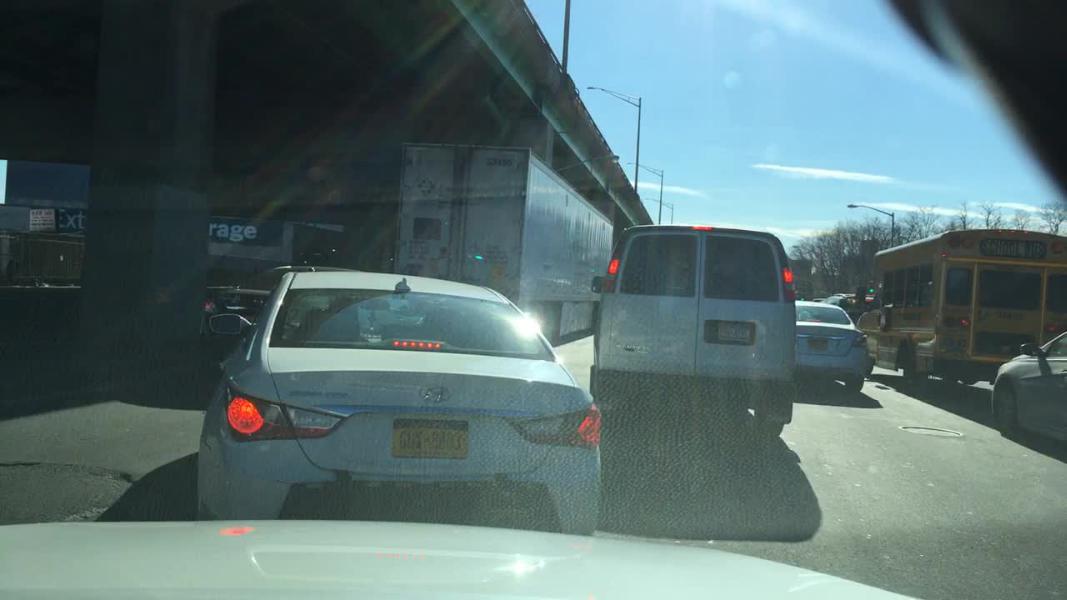}
    \end{minipage}
    \begin{minipage}{0.24\linewidth}
    \includegraphics[width=22mm,height=12mm]{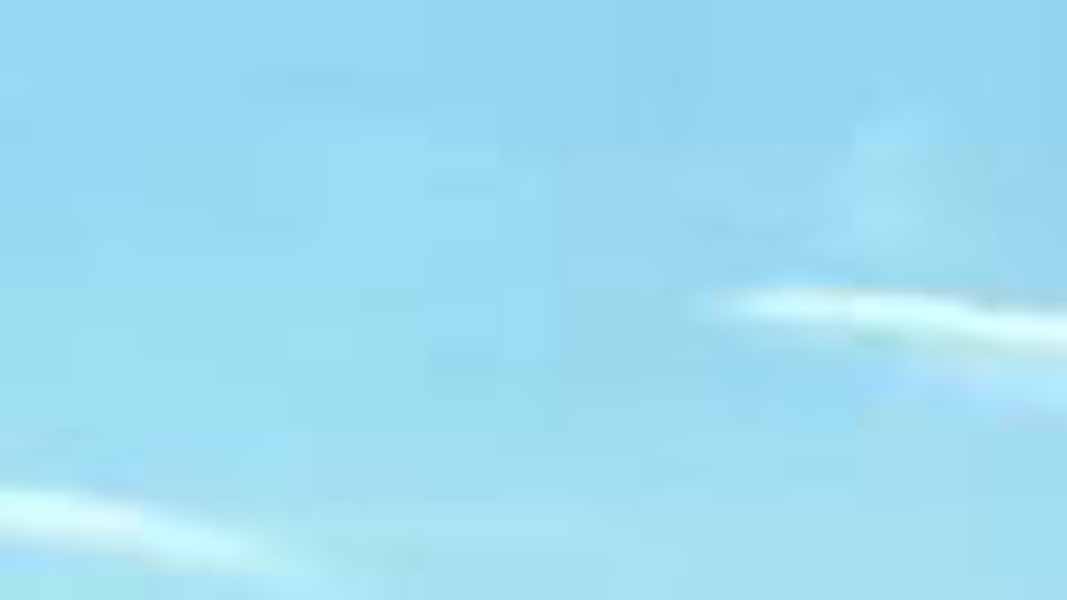}
    \end{minipage}
    \begin{minipage}{0.24\linewidth}
    \includegraphics[width=22mm,height=12mm]{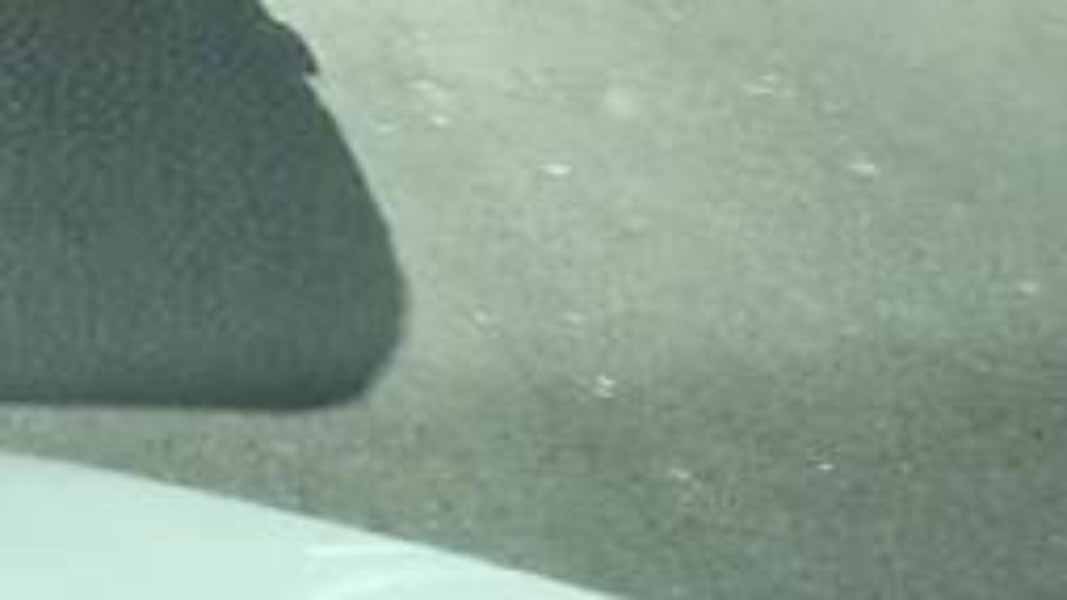}
    \end{minipage}
    \begin{minipage}{0.24\linewidth}
    \includegraphics[width=22mm,height=12mm]{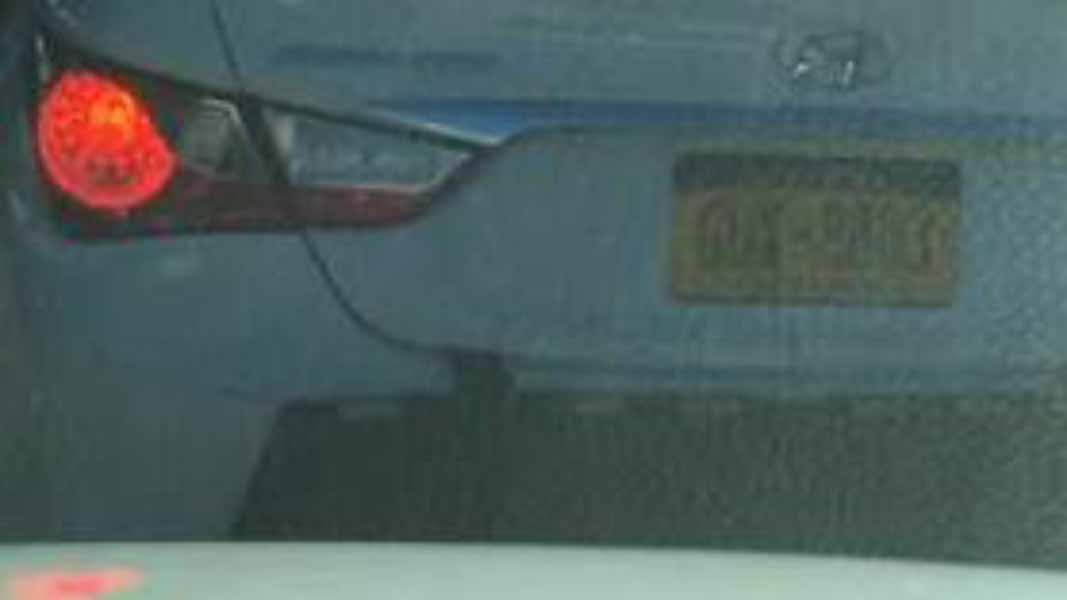}
    \end{minipage}

    \vspace{0.3mm}
    \begin{minipage}{0.24\linewidth}
    \includegraphics[width=22mm,height=12mm]{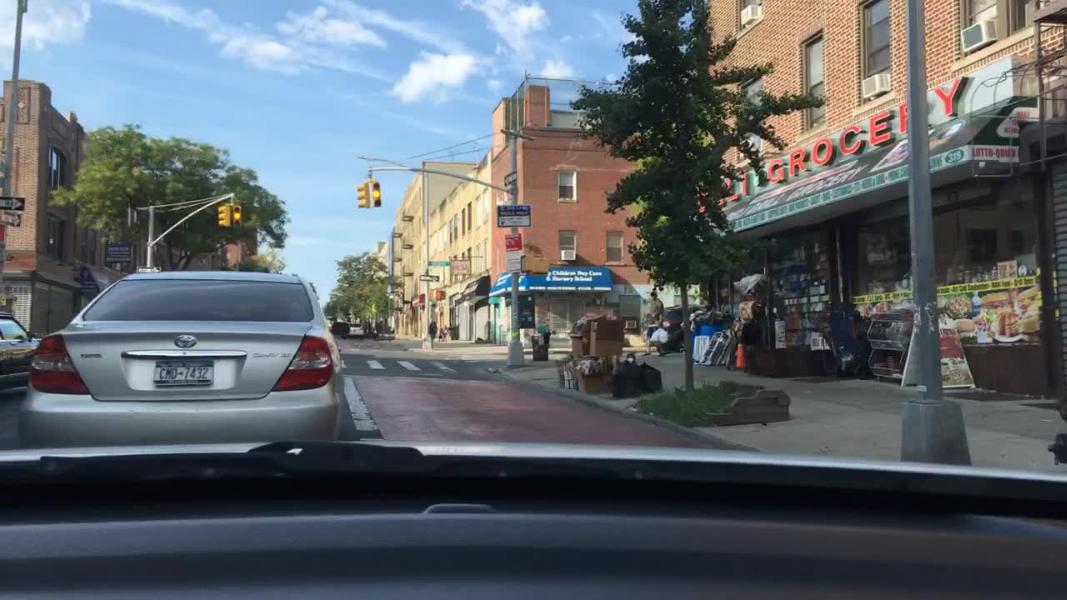}
    
    \centering
    \end{minipage}
    \begin{minipage}{0.24\linewidth}
    \includegraphics[width=22mm,height=12mm]{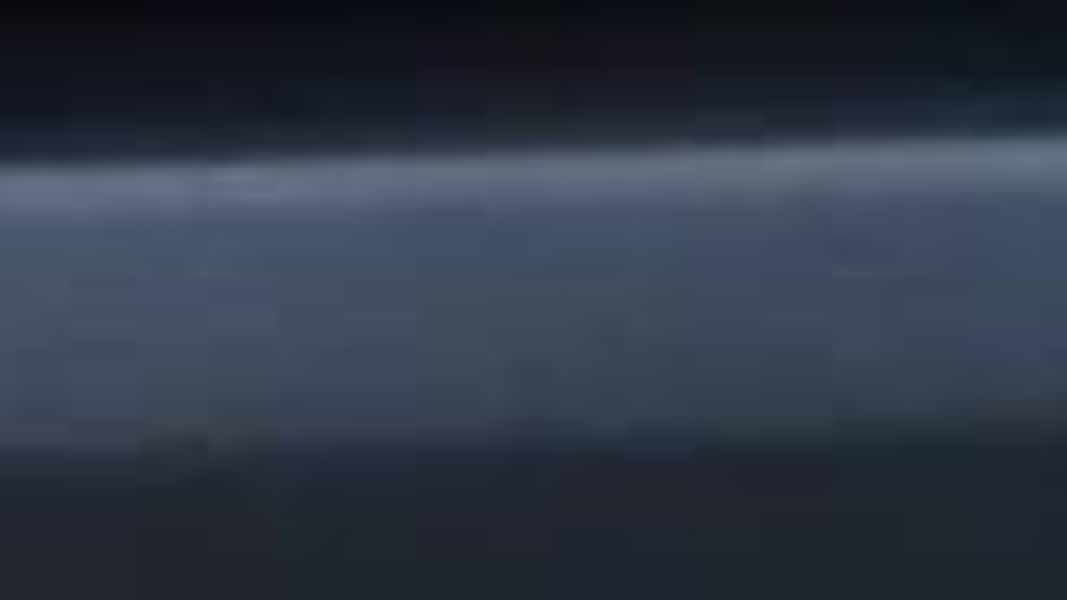}
    
    \centering
    \end{minipage}
    \begin{minipage}{0.24\linewidth}
    \includegraphics[width=22mm,height=12mm]{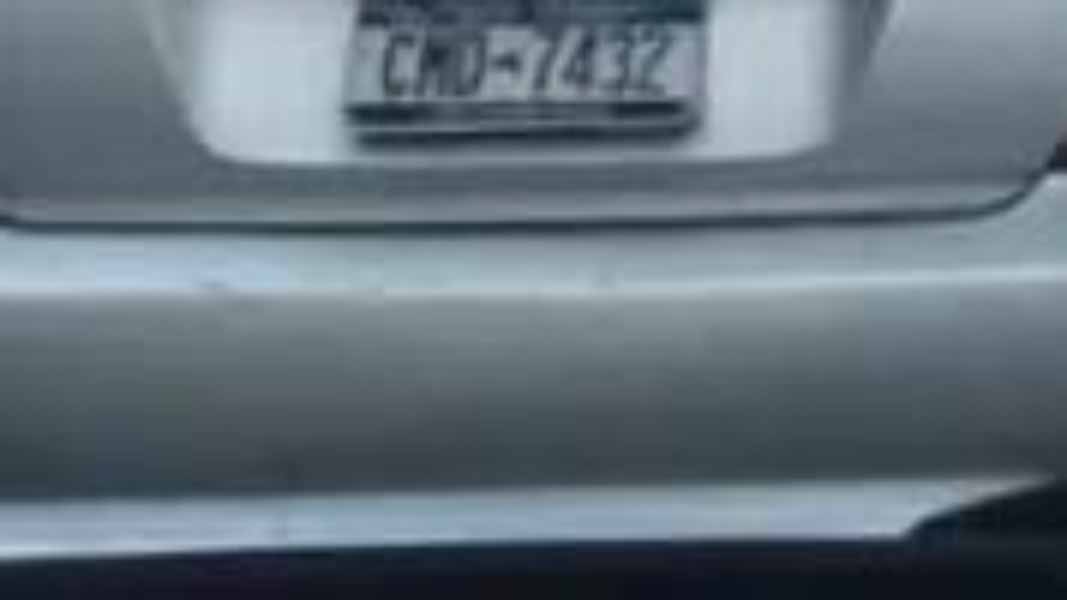}
    
    \centering
    \end{minipage}
    \begin{minipage}{0.24\linewidth}
    \includegraphics[width=22mm,height=12mm]{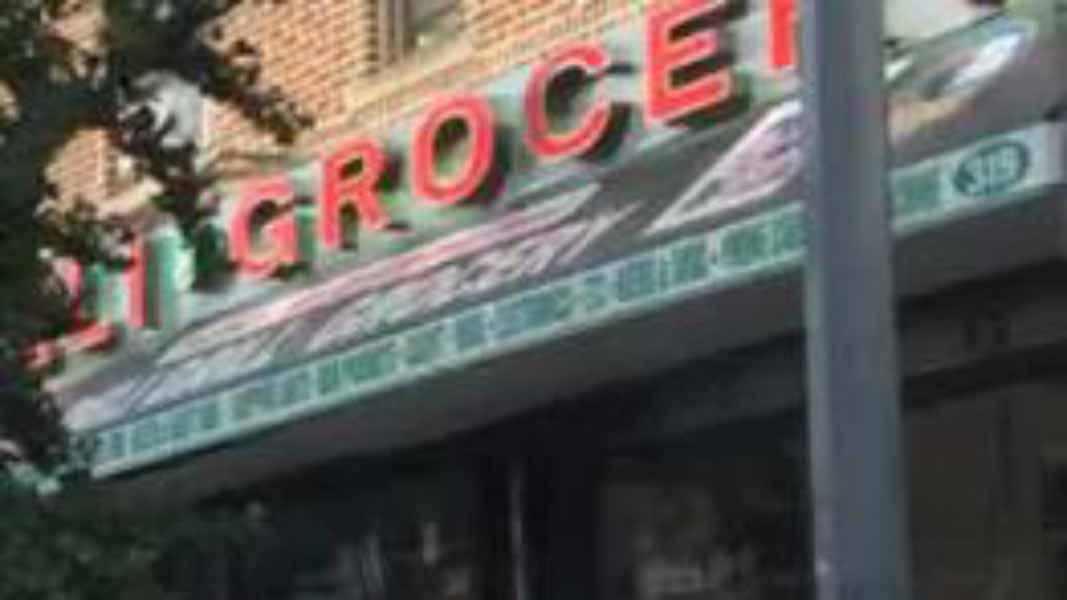}
    
    \centering
    \end{minipage}
    
    \caption{Visualization of patch selection. \textbf{First column}: original image. \textbf{Second and third columns}: discarded patch. \textbf{Fourth column}: selected patch.}
    \label{fig:glcm}
    \vspace{5mm}
    \end{minipage}
\end{figure}

\noindent\textbf{Visualization of patch selection.} We performed a visual analysis of the TBSA module, and the results are shown in Fig.~\ref{fig:glcm}. We can see that the texture of the discarded patches is relatively smooth and the pattern lacks variation. In addition, it can be observed that elements such as sky and road occupy most of the area of the image and are more likely to be selected randomly. Therefore, the GLCM-based filtering mechanism is reasonable.

\noindent\textbf{Qualitative detection results.} Fig.~\ref{fig:visbox} demonstrates some detection results under different weather conditions. Compared with Faster-RCNN, our method reduces the chance of wrong and missed detections. For example, the baseline method treats the near light as the foreground (Third column) and the distant vehicles as the background (Second column). For small objects in the distance and some blurred objects, our method performs a more accurate detection. It demonstrates that our approach improves the model's resistance to irrelevant attributes, resulting in better cross-domain detection performance.

\section{Conclusion}

In this work, we delve into the problem of model generalization under the single source domain for object detection. We analyze that existing methods lack the exploration of two key issues, first, the pseudo-correlation link between irrelevant attributes and labels, and second, the semantic structural relationships between samples that can facilitate the model's generalization ability. To this end, we propose a novel framework named SRCD, consisting of two key components, TBSA and LGSR. TBSA exploits the characteristic that the magnitude spectrum carries more domain-relevant information to change the image style, forcing the model to focus on domain-invariant information. Based on the stochasticity of TBSA, the single-source domain is transformed into the compound domains. LGSR aims to uncover and learn the semantic relationships among samples from the compound domains and empower the model to reason by maintaining a robust semantic structure, thus enhancing the model's ability to generalize across domains. Experiments on multiple benchmarks demonstrate the effectiveness of our approach.

\bibliographystyle{IEEEtran}
\bibliography{ref.bib}

\end{document}